\documentclass{article}

\usepackage{hyperref}
\hypersetup{pagebackref=true,breaklinks=true,colorlinks,bookmarks=false}  %
\usepackage{authblk}                %
\usepackage[symbol]{footmisc}       %
\usepackage{microtype}              %

\usepackage{bm}
\usepackage{amssymb}
\usepackage{amsmath}

\usepackage[binary-units=true]{siunitx}
\DeclareSIUnit\px{px}
\DeclareSIUnit\wattpeak{W_{P}}

\usepackage{subcaption}

\usepackage{cleveref}
\crefname{section}{Sec.}{Sections}
\crefname{figure}{Fig.}{Figure}
\crefname{table}{Tab.}{Table}
\crefname{equation}{Eq.}{Equation}
\crefname{appendix}{Appendix}{Appendix}

\usepackage{xspace}
\makeatletter
\DeclareRobustCommand\onedot{\futurelet\@let@token\@onedot}
\def\@onedot{\ifx\@let@token.\else.\null\fi\xspace}
\makeatother
\newcommand{\etal}[1]{#1~\textit{et~al\onedot}}

\newcommand{\eg}{e.\,g.,\xspace}

\newcommand{\cf}{cf\onedot}
\newcommand{\ie}{i.\,e.,\xspace}
\newcommand{\wrt}{w.\,r.\,t\onedot}

\usepackage{booktabs}

\usepackage{etoolbox} %
\usepackage{acronym}
\input{accaps.tex}

\newacro{PV}{photovoltaic}
\newacro{EL}{electroluminescense}
\newacro{IR}{infrared}
\newacro{IoU}{intersection over union}
\newacro{SVR}{support vector regression}
\newacro{DL}{deep-learning}
\newacro{MAE}{mean absolute error}
\newacro{MSE}{mean squared error}
\newacro{RMSE}{root mean squared error}
\newacro{MP}[PMPP]{power at maximum power point}
\newacro{MPP}{maximum power point}
\newacro{DNN}{deep neural network}
\newacro{SGD}{stochastic gradient descent}
\newacro{FC}{fully connected}
\newacro{CAM}{class activation map}
\newacro{GAP}{global average pooling}
\newacro{FCL}{fully connected layer}
\newacro{ZCA}{zero component analysis}
\newacro{t-SNE}{t-stochastic neighbour embedding}
\newacro{CV}{cross validation}
\newacro{STC}{standard test conditions}

\usepackage[inline]{enumitem}
\setlist*[enumerate]{label=(\arabic*)}

\usepackage{csvsimple}

\usepackage[page]{appendix}

\newcommand{\addition}[2][1=1]{#2}
\newcommand{\deletion}[2][1=1]{}
\newcommand{\change}[3][1=1]{#3}

\newcommand{\pytorch}{PyTorch~v1.3~\cite{paszke2019pytorch}\xspace}
\newcommand{\sklearn}{Scikit-learn library~v0.20~\cite{pedregosa2011scikit}\xspace}
\newcommand{\optuna}{Optuna~library~v1.2~\cite{akiba2019optuna}\xspace}

\newcommand{\myvector}[1]{\mathrm{#1}}

\newcommand{\typet}{T1\xspace}
\newcommand{\typee}{T2\xspace}
\newcommand{\typeh}{T3\xspace}

\newcommand{\resneta}{ResNet18\xspace}
\newcommand{\resnetb}{ResNet50\xspace}
\newcommand{\mobilenet}{MobileNetV2\xspace}

\newcommand{\mysample}{\ensuremath{\myvector{x}}\xspace}

\newcommand{\fresnet}{\ensuremath{\myvector{f}_{\text{R18}}}\xspace}
\newcommand{\fmstd}{\ensuremath{\myvector{f}_{\mu,\sigma}}\xspace}

\newcommand{\fresnetact}{\ensuremath{\myvector{f}}\xspace}
\newcommand{\fcam}{\ensuremath{\myvector{{f}_{\text{cam}}}}\xspace}

\newcommand{\pnom}{\ensuremath{P_{\text{nom}}}\xspace}
\newcommand{\prel}{\ensuremath{p_{\text{rel}}}\xspace}
\newcommand{\prelest}{\ensuremath{\hat{p}_{\text{rel}}}\xspace}
\newcommand{\perrrel}{\ensuremath{\Delta p_{\text{rel}}}\xspace}

\newcommand{\mae}{\ensuremath{\text{MAE}}\xspace}
\newcommand{\rmse}{\ensuremath{\text{RMSE}}\xspace}

\newcommand{\pmpp}{\ensuremath{P_{\text{mpp}}}\xspace}
\newcommand{\pmppest}{\ensuremath{\hat{P}_{\text{mpp}}}\xspace}
\newcommand{\nsamples}{\ensuremath{N}}

\newcommand{\mymean}{\ensuremath{\mu}\xspace}
\newcommand{\mystd}{\ensuremath{\sigma}\xspace}
\newcommand{\mylr}{\ensuremath{\eta}\xspace}

\newcommand{\mybs}{\ensuremath{B}\xspace}
\newcommand{\mymomentum}{\ensuremath{\nu}\xspace}
\newcommand{\mywd}{\ensuremath{\lambda}\xspace}

\newcommand{\prescale}{\ensuremath{s}\xspace}           %
\newcommand{\pretsize}{\ensuremath{\tau_s}\xspace}      %
\newcommand{\pretiou}{\ensuremath{\tau_{\text{IoU}}}\xspace}

\newcommand{\resultsmaew}[1]{\csvloop{
                    file=csv_results/039_summary/summary.csv,
                    head to column names,
                    filter strcmp={\model}{#1},
                    command={\SI[separate-uncertainty=true, multi-part-units=single]{\maeW(\uncertaintyW)}{\wattpeak}},
                }%
}

\newcommand{\resultsmae}[1]{\csvloop{
                    file=csv_results/039_summary/summary.csv,
                    head to column names,
                    filter strcmp={\model}{#1},
                    command={\SI[separate-uncertainty=true, multi-part-units=single]{\mae(\uncertainty)}{\percent}},
                }%
}
\newcommand{\resultsgeneralizationmae}[1]{\csvloop{
                    file=csv_results/041_check_generalization/summary.csv,
                    head to column names,
                    filter strcmp={\model}{#1},
                    command={\SI[separate-uncertainty=true, multi-part-units=single]{\mae(\uncertainty)}{\percent}},
                }%
}

\usepackage{tikz}
\usetikzlibrary{patterns}
\usetikzlibrary{calc}
\usetikzlibrary{positioning,fit}
\usetikzlibrary{arrows.meta}
\usepackage{rotating}
\usepackage{verbatim}
\usepackage{pgfplots}
\pgfplotsset{compat=1.14}
\usepgfplotslibrary{statistics}
\usepackage{pgfplotstable}
\usepgfplotslibrary{external,patchplots}
\usepgfplotslibrary{colorbrewer,groupplots}
\pgfdeclarelayer{bg}
\pgfsetlayers{bg,main}  %

\pgfplotsset{
    discard if/.style 2 args={
        x filter/.append code={
            \edef\tempa{\thisrow{#1}}
            \edef\tempb{#2}
            \ifx\tempa\tempb
                
            \fi
        }
    },
    discard if not/.style 2 args={
        x filter/.append code={
            \edef\tempa{\thisrow{#1}}
            \edef\tempb{#2}
            \ifx\tempa\tempb
            \else
                
            \fi
        }
    },
    discard boxplot if not/.style 2 args={
        /pgfplots/boxplot/data filter/.code={
            \edef\tempa{\thisrow{#1}}
            \edef\tempb{#2}
            \ifx\tempa\tempb
            \else
                
            \fi
        }
    }
}

\definecolor{colorh2}{RGB}{27,158,119}
\definecolor{colort1}{RGB}{217,95,2}
\definecolor{colore3}{RGB}{117,112,179}

\pgfplotsset{
    scatterclasses style/.style={
        scatter/classes={
            t_high={mark=o,colort1},
            lzv_s_high={mark=triangle,colort1,scale=2.0},
            lzv_a_high={mark=triangle,colort1,scale=2.0},
            h_high={mark=o,colorh2},
            h_low={mark=*,colorh2},
            lzv_e19_high={mark=triangle,colore3,scale=2.0},
            lzv_e19_low={mark=triangle*,colore3,scale=2.0}
        },
    },
}
\pgfplotsset{
    scatterplotnew style/.style={
        enlargelimits=false,
        axis on top,
        xlabel={\pmpp [\si{\wattpeak}]},
        ylabel={\pmppest [\si{\wattpeak}]},
        scatterclasses style,
        xmin=80,
        xmax=280,
        ymin=80,
        ymax=280,
    },
}
\newcommand{\scattercommon}{
    \addplot [
        domain=0:300,
        samples=2,
        no markers,
    ] {x};
    \addplot [
        domain=0:300,
        samples=2,
        no markers,
        dashed,
        white!50!black,
    ] {x+15};
    \addplot [
        domain=0:300,
        samples=2,
        no markers,
        dashed,
        white!50!black,
    ] {x-15};
}

\pgfplotsset{
    scatter samples/.style={
        only marks,
        scatter,
        scatter src=explicit symbolic,
        mark options={scale=0.5},
    }
}

\usepackage{textcmds}

\begin{document}

\suppressfloats

\title{Deep Learning-based Pipeline for Module Power Prediction from EL Measurements}

\author[1,3]{Mathis Hoffmann$^{*}$}
\author[2]{Claudia Buerhop-Lutz$^{*}$}
\author[1,2]{Luca Reeb}
\author[2]{Tobias Pickel}
\author[3,2]{Thilo Winkler}
\author[2,3,4]{Bernd Doll}
\author[1]{Tobias W\"urfl}
\author[2]{Ian Marius Peters}
\author[2,3,4]{Christoph Brabec}
\author[1,4]{Andreas Maier}
\author[1]{Vincent Christlein}

\affil[1]{Pattern Recognition Lab, Universit\"at Erlangen-N\"urnberg (FAU)}
\affil[2]{Forschungszentrum J\"ulich GmbH, Helmholtz-Institute Erlangen-Nuremberg for Renewable Energies}
\affil[3]{Institute Materials for Electronics and Energy Technology, FAU}
\affil[4]{School of Advanced Optical Technologies, Erlangen}

\date{\vspace{-5ex}}    %

\maketitle

\footnotetext[1]{equal contribution}

\begin{abstract}
Automated inspection plays an important role in monitoring large-scale photovoltaic power plants. Commonly, \acl{EL} measurements are used to identify various types of defects on solar modules, but have not been used to determine the power of a module. However, knowledge of the \acl{MP} is important as well, since drops in the power of a single module can affect the performance of an entire string. By now, this is commonly determined by measurements that require to discontact or even dismount the module, rendering a regular inspection of individual modules infeasible. In this work, we bridge the gap between \acl{EL} measurements and the power determination of a module. We compile a large dataset of \num{719} \acl{EL} measurements of modules at various stages of degradation, especially cell cracks and fractures, and the corresponding \acl{MP}. Here, we focus on inactive regions and cracks as the predominant type of defect. We set up a baseline regression model to predict the power from \acl{EL} measurements with a \acl{MAE} of \resultsmaew{svrmstd} (\resultsmae{svrmstd}). Then, we show that \acl{DL} can be used to train a model that performs significantly better (\resultsmaew{resnet18} or \resultsmae{resnet18}) and propose a variant of class activation maps to obtain the per cell power loss, as predicted by the model. With this work, we aim to open a new research topic. Therefore, we publicly release the dataset, the code and trained models to empower other researchers to compare against our results. Finally, we present a thorough evaluation of certain boundary conditions like the dataset size and an automated preprocessing pipeline for on-site measurements showing multiple modules at once.

\end{abstract}

\acresetall

\section{Introduction}

\begin{figure}
    \centering
    \tikzsetnextfilename{overview}
    \begin{tikzpicture}[
        node distance=1.5cm
    ]

        \tikzstyle{nnet}=[
            circle,
            fill=Paired-B,
            minimum width=0.3cm,
            anchor=center,
            node distance=0.25cm
        ]
        \tikzstyle{nextarrow}=[
            -Latex,
            thick,
        ]
        \tikzstyle{box}=[
            minimum height=2.7cm,
            fill=Paired-A,
            rounded corners=0.3cm,
            inner sep=0.3cm,
            outer sep=0.2cm,
        ]

        \node[inner sep=0pt](input) at (0,0)
            {\includegraphics[width=2cm]
            {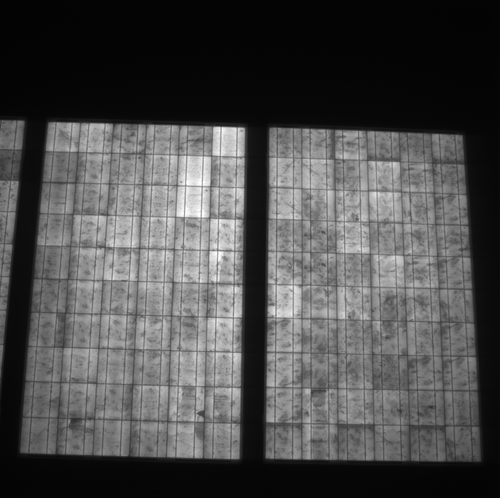}};

        \node[anchor=center, fill=Paired-B, right=of input](preproc) {\color{white}\huge$\circlearrowleft$};
        \draw[nextarrow] ($(input.east)!0.25!(preproc.west)$)--($(input.east)!0.75!(preproc.west)$);

        \coordinate[right=0.5cm of preproc](result);
        \draw[->] (preproc.east)--(result);
        \node[inner sep=0pt, anchor=west](res1) at ($(result) + (0,0.1)$)
            {\includegraphics[width=1.5cm]
            {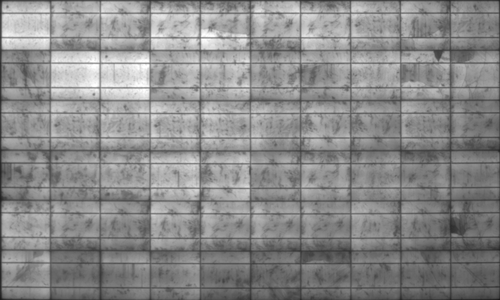}};
        \node[inner sep=0pt, anchor=west](res2) at ($(result) + (0.1,-0.1)$)
            {\includegraphics[width=1.5cm]
            {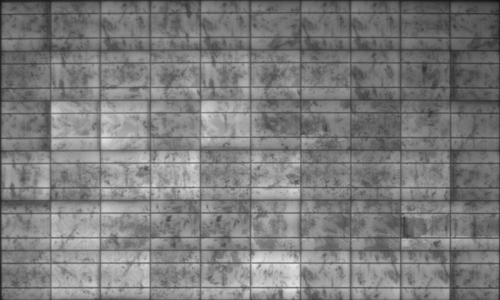}};
        \coordinate[](resultright) at ($(res2.east)+(0,0.1)$);
        
        \coordinate[right= of resultright](netbegin);
        \node[nnet](x1) at (netbegin) {};
        \node[nnet, above=of x1](x0) {};
        \node[nnet, below=of x1](x2) {};
        \coordinate(x01) at ($(x0)!0.5!(x1)$);
        \draw[nextarrow] ($(resultright)!0.25!(x1.west)$)--($(resultright)!0.75!(x1.west)$);

        \node[nnet, right=of x01](y0) {};
        \node[nnet, below=of y0](y1) {};
        \coordinate(y01) at ($(y0)!0.5!(y1)$);
        \draw[] (x0)--(y0);
        \draw[] (x1)--(y0);
        \draw[] (x2)--(y0);
        \draw[] (x0)--(y1);
        \draw[] (x1)--(y1);
        \draw[] (x2)--(y1);

        \node[nnet, right=of y01](z0) {};
        \draw[] (y0)--(z0);
        \draw[] (y1)--(z0);

        \node[right=0.5cm of z0](presult) {\prelest};
        \draw[->] (z0)--(presult);

        \begin{pgfonlayer}{bg}
            \node[box, fit=(input), label=above:{\cref{sec:data}}](box1){};
            \node[box, fit=(preproc)(res1)(res2), label=above:{\cref{sec:localization}}](box2){};
            \node[box, fit=(x0)(x2)(presult), label=above:{\cref{sec:powerprediction}}](box3){};
        \end{pgfonlayer}
        
    \end{tikzpicture}
    \caption{Overview of this work. First, data is collected on-site as well as under lab conditions. This is detailed in~\cref{sec:data}. Then, measurements are preprocessed to obtain a sequence of single modules, as described in~\cref{sec:localization}. Here, preprocessing is denoted as $\circlearrowleft$. Finally, power prediction is applied to obtain the estimated module power $\prelest$ relative to the nominal power $\pnom$. This is summarized by~\cref{sec:powerprediction}.}
    \label{fig:overview}
\end{figure}
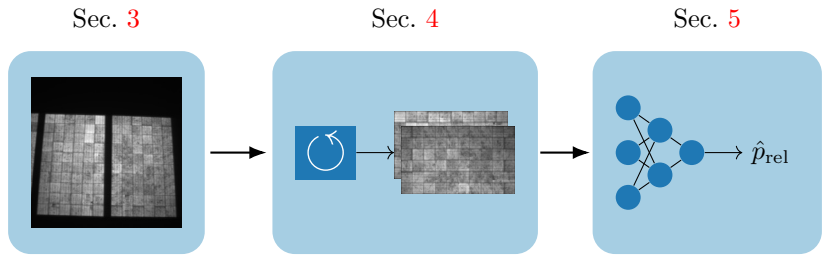

Over the last years, \ac{PV} power production has become an important factor in the energy production worldwide. This is promoted by decreasing module cost, increasing module power, long lifetime and low maintenance cost. More important, it produces power at a low ecological footprint and is hence used to fight global heating. However, continuous monitoring or regular inspections of \ac{PV} power plants are necessary to ensure a constant and safe operation. In addition, modules might be damaged during manufacturing, transport, installation or operation and an early detection of those cases can help to avoid a later replacement at high cost.

It is common practice to operate multiple \ac{PV} modules connected in series, commonly referred to as strings. As a result, a failure of a single module can drastically degrade the performance of the entire string. However, continuous monitoring is usually applied to strings and not to individual modules for economical reasons. To this end, manual inspection of single modules is required to further narrow down failures.

Lately, \ac{EL} imaging is accepted as a useful tool by the community to analyze many failures of single modules, \eg cell cracks and fractures. This is because many defect types can be identified easily and, as opposed to \ac{IR} measurements, a quantification of the active area of a cell is possible~\cite{buerhop2018evolution}. The latter makes it especially well suited for automated prediction of the module power by statistical methods like \ac{DL}. Traditionally, the \ac{MPP} is determined from direct measurements of the IV curve. Although this is possible on-site, it requires to disconnect every single module to perform the measurement. Hence, this is time consuming and costly. \addition[ref=co025,label=ch014]{In contrast, \ac{EL}-based determination of the \ac{MPP} enables an automated inspection of entire power plants, possibly also using unmanned aerial vehicles.}

For \ac{PV}-plant operators especially cell cracks and fractures with varying origin, including manufacturing process, transport, installation, operating conditions (\eg storm, hailstorm, snow load) are of particular relevance. \addition[label=ch000,ref=co001]{Note that, throughout this work, we refer to breaks of the cell that cause parts of it to become disconnected as fractures, whereas cracks are breaks that do not lead to disconnected regions}. The impact of cell cracks on the module power and performance is studied intensely during the last years using EL-images for identifying cracks and IV-tracers for determining the module power~\cite{kontges2011crack,kajari2011spatial,sander2011investigations,paggi2016global,buerhop2018evolution,gabor2015designfactors}. To study crack propagation, mostly climate chambers and static loading, \eg using sand sacks, were used. Buerhop and Gabor studied the performance of modules during static and cyclic loading using a specialized setup~\cite{buerhop2018cycleperformance,gabor2016mechanical}, where \ac{EL}-measurements as well as power data were recorded at loaded and unloaded stages. In addition to the indoor experiments, Buerhop studied the performance of modules with cracked cells at an on-site test facility in detail~\cite{buerhop2017fieldperformance}. All previous investigations have in common that two separate measurements are required: The \ac{EL}-measurements are used to identify cracks and IV-tracing is used to determine the \ac{MP}. Given that on-site IV measurements are costly, an automated and reliable estimation of the \ac{MP} and the impact of cell cracks and fractures on the latter using \ac{EL}-measurements are of particular importance for future studies. Note that, throughout this work, we use the terms module power and \acl{MP} interchangeably.

In a previous conference paper, we showed that the \ac{MP} can be automatically predicted from \ac{EL} measurements \change[ref=co030,label=ch024]{measured indoors}{taken during indoor mechanical load testing experiments}~\cite{buerhop2019dlpower}. This work is a direct continuation and extension of that conference paper. As opposed to the previous work, we now focus on building regression models for on-site data. For efficiency reasons, on-site \ac{EL}-measurements are usually conducted such that multiple modules are visible in a single measurement. Since we aim to predict the \ac{MP} for single modules, we design a segmentation pipeline for on-site \ac{EL}-measurements to automatically generate images showing only a single module. Furthermore, we add an extensive evaluation of the prediction performance, stability and boundary conditions. Finally, we automatically quantify the per cell power loss. The main contributions of this work are as follows:
\begin{enumerate}
    \item We develop a method to predict the \ac{MP} from a single on-site or indoor \ac{EL} module measurement. We achieve a \ac{MAE} of \resultsmaew{resnet18} \deletion[ref=co006]{under the condition that all failure types that have an impact on the module power are visible in the \ac{EL} measurement} and compare the result against alternative methods in a thorough evaluation.\label{itm:contrib-pp} \addition[ref=co006,label=ch049]{Since the dataset is specifically selected such that modules have a high shunt resistance, the trained models are restricted to this defect type.}
    \item We propose a fast and robust pipeline to detect and segment multiple modules in \ac{EL} measurements, such that \cref{itm:contrib-pp} can be directly applied for on-site assessment of modules.\label{itm:contrib-preproc}
    \item We perform an extensive evaluation of boundary conditions, such as minimum image size and dataset size.\label{itm:contrib-cond}
    \item We set up and publicly release a dataset that is specifically compiled for the task at hand~\cite{juelich2020pvpowerdata}. In addition, we publish our code and trained models\footnote{\url{https://github.com/ma0ho/elpvpower}}, such that a direct application and comparison in future research is feasible and that our results directly scale to any other \ac{PV} plant.\label{itm:contrib-open}
    \item We predict the the per cell power loss and analyze the results in terms of defect severity. For our dataset that is compiled such that modules show a reduced active area in \ac{EL} measurements, we find that the \ac{MP} is dominated by fractures and that cracks are only of minor importance.
\end{enumerate}

The remainder of this work is organized as shown in~\cref{fig:overview}: In~\cref{sec:data}, we describe the data collection procedure and characterize the dataset used in this work. In~\cref{sec:localization}, we describe the localization of multiple modules from single \ac{EL} measurements. Then, in~\cref{sec:powerprediction}, we detail the automated power prediction. This includes baseline methods as well as \ac{DL}-based methods.

Since the different parts of this work are of interest independent of each other, we decided to summarize the results of every part directly after describing the methodology and omit a detailed global results section. Then, we finally summarize the most important results in~\cref{sec:conclusion}.

\section{Related Work}

In the last years, many efforts have been made to leverage computer vision methods in order to reduce the maintenance and operating cost of \ac{PV} power plants. This includes automated defect analysis for solar cells~\cite{stromer2019enhanced,mayr2019weakly,deitsch2019automatic} or automated prediction of solar irradiance~\cite{bernecker2014continuous}. \change[label=ch001,ref=co002]{To the best of our knowledge, MP characterization of a PV module using a single EL measurement has not been reported before}{In prior works, it has been shown that the number of cracks visible in \ac{EL} measurements roughly correlates to the power loss of a module~\cite{dubey2018site} and that the size of the inactive area on a module correlates well to the power of the module, as long as the overall fraction of inactive area remains small~\cite{schneller2018electroluminescence}. Furthermore, \etal{Karimi} have shown that the normalized power of the module can be determined from various features extracted from the \ac{EL} image~\cite{karimi2020generalized}}. In addition, there are a few works on \ac{MP} characterization using other methods as well. For example, \etal{Teubner} propose to use \ac{IR} measurements of a module affected by potential induced degradation and compute the \ac{MP} using linear regression from the mean temperature difference of the module to a reference module~\cite{teubner2019irpower2}. However, this procedure requires that the reference module is exposed to the same environmental conditions (air temperature, wind speed) as the module under test. This is especially challenging for roof-mounted \ac{PV} installations, where the a temperature gradient is present due to the convective environment. Furthermore, the measurement accuracy is highly dependent on the available measurement time and steady environmental conditions. Recently, \etal{Ortega} proposed a continuous monitoring of \ac{PV} power plants using measurement devices attached to every module~\cite{ortega2019pvpower}. \etal{Kropp} manage to predict cell level characteristics from two \ac{EL} measurements using simulations and finally calculate the module power for a single test module~\cite{kropp2018quantitative}.

In contrast to the \ac{IR}-based approach, our method does not require a reference temperature, since the magnitude of \ac{EL} measurements mainly depends on camera characteristics and the excitation current, which is known in advance. Further, it is based on automated detection of defective areas rather than using module-wide statistics only. In contrast to the measurement-based approach by \etal{Ortega}, no additional hardware needs to be attached to the modules and the type and location of a defect can be determined additionally. As opposed to the simulation-based approach by \etal{Kropp}, our method only requires a single \ac{EL} measurement of a module. \addition[ref=co002,label=ch025]{In comparison to previous works on power estimation using a single \ac{EL} measurement, our method does not rely on hand-crafted features. Instead, relevant features are learned from the data directly, resulting in a data-optimal regression model. Since no manual feature design is required, this method generalizes to various defect and module types, given an appropriate training dataset. Furthermore, it enables the visualization of the learned features and quantification of per cell or even per defect power losses, facilitating a better understanding for the roots of power degradation.}

\section{Data}\label{sec:data}

\begin{figure*}
    \begin{subfigure}{.33\textwidth}
        \includegraphics[width=\linewidth]{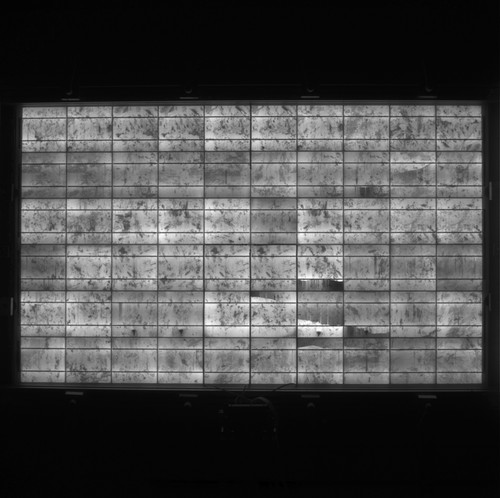}
        \caption{\typet (indoor/on-site)}
        \label{fig:data-t}
    \end{subfigure}%
    \hfill%
    \begin{subfigure}{.33\textwidth}
        \includegraphics[width=\linewidth]{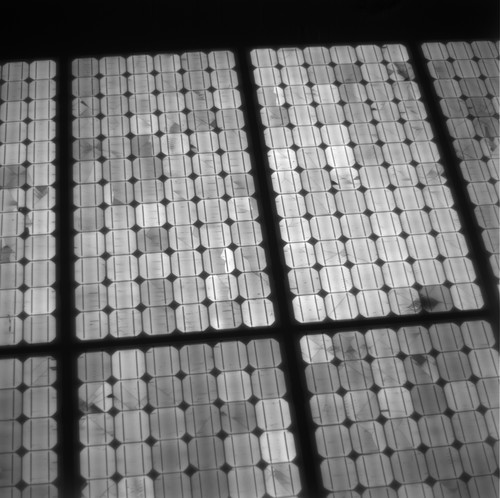}
        \caption{\typee (on-site)}
        \label{fig:data-e}
    \end{subfigure}%
    \hfill%
    \begin{subfigure}{.33\textwidth}
        \includegraphics[width=\linewidth]{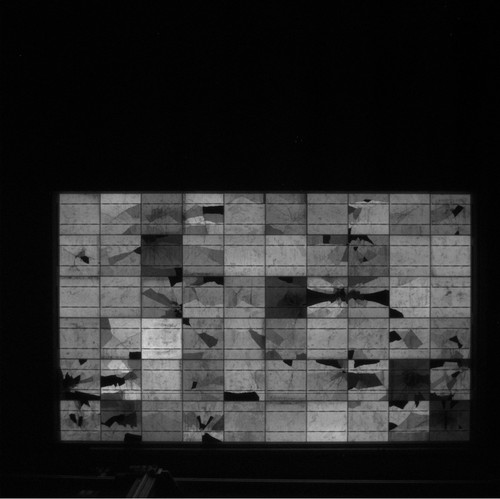}
        \caption{\typeh (indoor)}
        \label{fig:data-h}
    \end{subfigure}%
    \caption{Example images from the data set.}
    \label{fig:data-examples}
\end{figure*}

\begin{figure}[t]
    \centering
    \tikzsetnextfilename{explore_data_datasets}
    \begin{tikzpicture}
        \begin{axis}[
                xbar stacked,
                bar width=0.1,
                width=0.6\textwidth,
                height=3cm,
                xmin=0,
                ymin=0.9,
                ymax=1.1,
                yticklabels={},
                ytick={},
                legend pos=outer north east,
                legend columns=2,
                xlabel={\nsamples},
            ]
            \addplot[draw=colort1, fill=colort1!20!white] table [y expr=1, x=count, discard if not={key}{t_high}, col sep=comma]{csv_results/005_explore_data/counts_by_moduletype_and_current.csv};
            \addplot[draw=colore3, fill=colore3!20!white] table [y expr=1, x=count, discard if not={key}{e_high}, col sep=comma]{csv_results/005_explore_data/counts_by_moduletype_and_current.csv};
            \addplot[draw=colore3, fill=colore3] table [y expr=1, x=count, discard if not={key}{e_low}, col sep=comma]{csv_results/005_explore_data/counts_by_moduletype_and_current.csv};
            \addplot[draw=colorh2, fill=colorh2!20!white] table [y expr=1, x=count, discard if not={key}{h_high}, col sep=comma]{csv_results/005_explore_data/counts_by_moduletype_and_current.csv};
            \addplot[draw=colorh2, fill=colorh2] table [y expr=1, x=count, discard if not={key}{h_low}, col sep=comma]{csv_results/005_explore_data/counts_by_moduletype_and_current.csv};
            
            \legend{\typet (high),\typee (high),\typee (low),\typeh (high),\typeh (low)};
        \end{axis}
    \end{tikzpicture}
    \caption{The dataset used in this work consists of three different module types (\typet, \typeh, \typee). Two of those have been measured at a high and low current, which indicated in brackets. Overall, the dataset consists of $\nsamples=719$ samples.}
    \label{fig:data-samplecounts}
\end{figure}
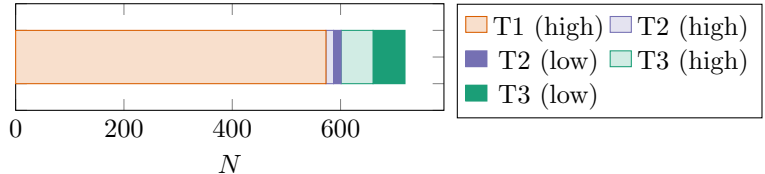

We collected a large set of \num{719} \ac{EL} measurements \addition[ref=co039,label=ch030]{showing \num{137} module instances} along with measurements of the maximum power point \pmpp \addition[ref=co031,label=ch047]{at standard test conditions (\SI{1000}{\watt\per\square\metre}, \SI{25}{\degree})}. The dataset is designed such that it has variations in the measurement procedure as well as in the type of solar modules. Overall, it contains three different module types, denoted as \typet, \typeh and \typee \addition[ref=co039,label=ch029]{with nominal powers given by \SIrange{225}{235}{\wattpeak}, \SIrange{235}{245}{\wattpeak} and \SIrange{170}{180}{\wattpeak} respectively}. Furthermore, it also includes variations in the excitation currents, such that models trained using the data should be invariant to the excitation current to some degree. \addition[ref=co035,label=ch027]{In particular, the excitation current for \typeh is given by $90\%_{ISC}$ (high current) and $10\%_{ISC}$ (low current), whereas for \typee it is given by $75\%_{ISC}$ (high current) and $50\%_{ISC}$ (low current).} Finally it includes on-site measurements as well as measurements taken at controlled indoor conditions. Examples from the dataset are shown in~\cref{fig:data-examples}. The distribution of samples between module types is shown in~\cref{fig:data-samplecounts}.

The EL measurements were recorded by a \qq{Greateyes 2048 2048} silicon detector camera with \SI{50}{\milli\metre} focal length lens and a camera triggered power supply. The camera parameters were fixed for the mechanical load testing site to an integration time of \SI{5}{\second} and an aperture of \num{2.4}. \deletion[ref=co004,label=ch003]{For on-site measurements, only the aperture was adjusted after the ambient conditions.}\addition[ref=co004]{They measurements have been recorded at 16-bit without image compression.} \addition[ref=co025,label=ch012]{Furthermore, the on-site measurements have been taken during night.} \addition[ref=co038,label=ch028]{Here, the integration time was also fixed to \SI{5}{\second} and the aperture was constant as well.}

The \ac{MP} of each PV module was determined from IV-curve measurements. For indoor measurements a table flasher \qq{Spire Spi-Sun Simulator 4600 SLP} with estimated measurement uncertainty of \SI{1.45}{\percent} was used, while the indoor measurements at the mechanical load testing site have been carried out using a prototype of a permanent light source consisting of many halogen lamps. The results were extrapolated to \ac{STC} with a measurement uncertainty of \SIrange{3}{4}{\percent}. For on-site measurements, a \qq{PVPM 1000CX} with a measurement uncertainty of at least \SI{5}{\percent} according to the data sheet, was used.

As seen from~\cref{fig:data-samplecounts}, the dataset is dominated by module type~\typet, because this type has been used for mechanical load testing~\cite{buerhop2017mechanicalload}. During that procedure, load is simulated by an underpressure that is applied on the backside of each module. The underpressure is increased stepwise. After every increase, an \ac{EL} and \ac{MP} measurement is taken under load. Then, pressure is released and another set of measurements is taken, before a new load cycle is started. As a result, there are about \num{50} sets of measurements of a single module at different stages of loading including changes in the crack structures. Since these different stages of degradation come with a variance in \ac{MP}, these load cycles are useful to assess, if a certain type of defect has an influence on the \ac{MP}. Furthermore, as shown in~\cref{fig:data-examples}, there is a series of on-site measurements using module type~\typet as well.

The set of measurements using module type~\typee only consists of on-site measurements. As opposed to \typet, we vary the \ac{EL} excitation current for this measurement series between high and low excitation. This is later used to obtain a model that \change[ref=co006,label=ch040]{is invariant to the excitation current}{performs well irrespective of the excitation current}. The same holds for module type~\typeh. This module type has been measured under indoor conditions again and includes variations in the excitation current, too. \addition[ref=co044,label=ch038]{However, this type has not been subject to load testing experiments. Instead, the degradation is caused by natural events like hailstorms.} An even more detailed analysis of the data can be found in~\cref{app:data-analysis}.

\section{Localization of multiple modules}\label{sec:localization}

\begin{figure*}
    \begin{subfigure}{.24\textwidth}
        \includegraphics[width=\linewidth]{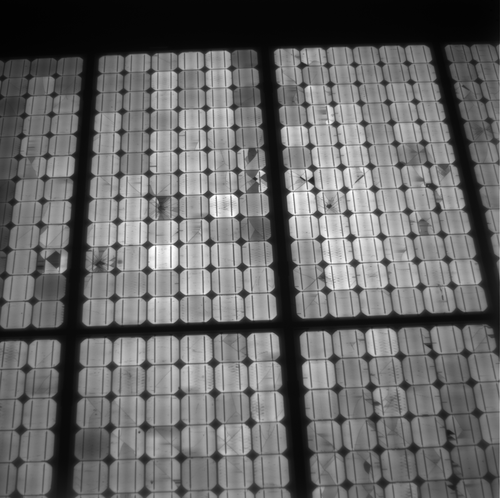}
    \end{subfigure}%
    \hfill%
    \begin{subfigure}{.24\textwidth}
        \includegraphics[width=\linewidth]{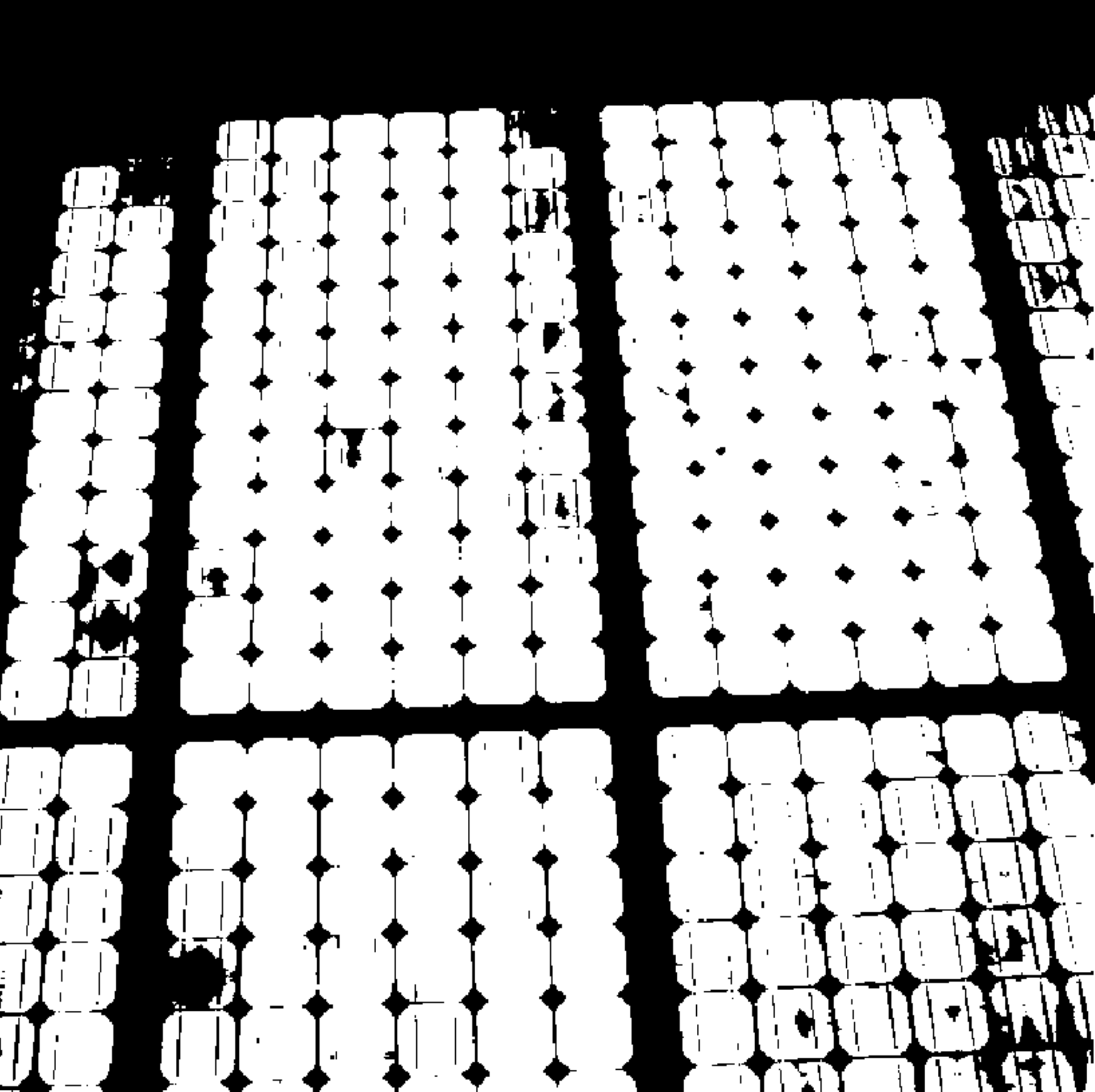}
    \end{subfigure}%
    \hfill%
    \begin{subfigure}{.24\textwidth}
        \includegraphics[width=\linewidth]{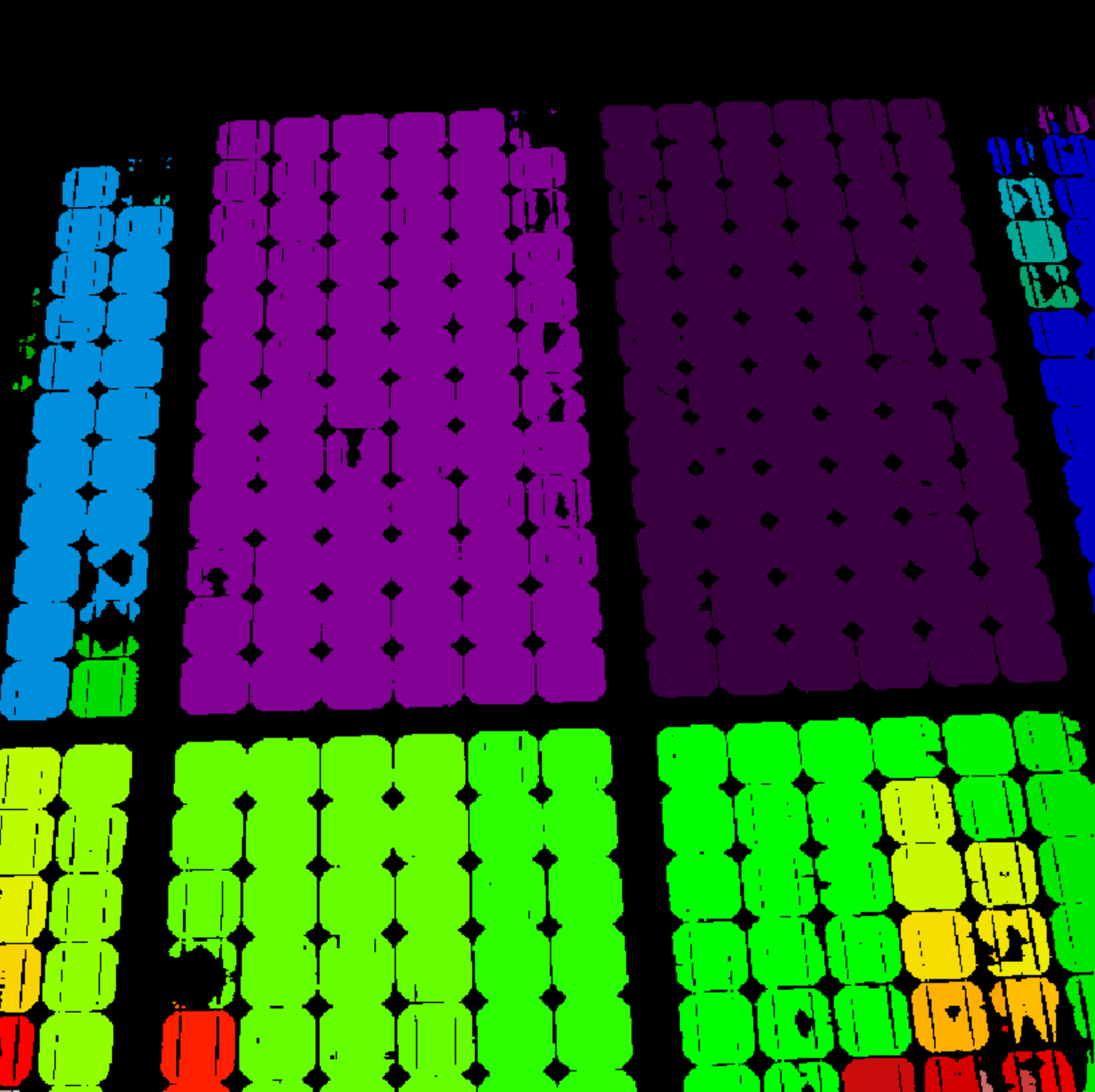}
    \end{subfigure}%
    \hfill%
    \begin{subfigure}{.24\textwidth}
        \includegraphics[width=\linewidth]{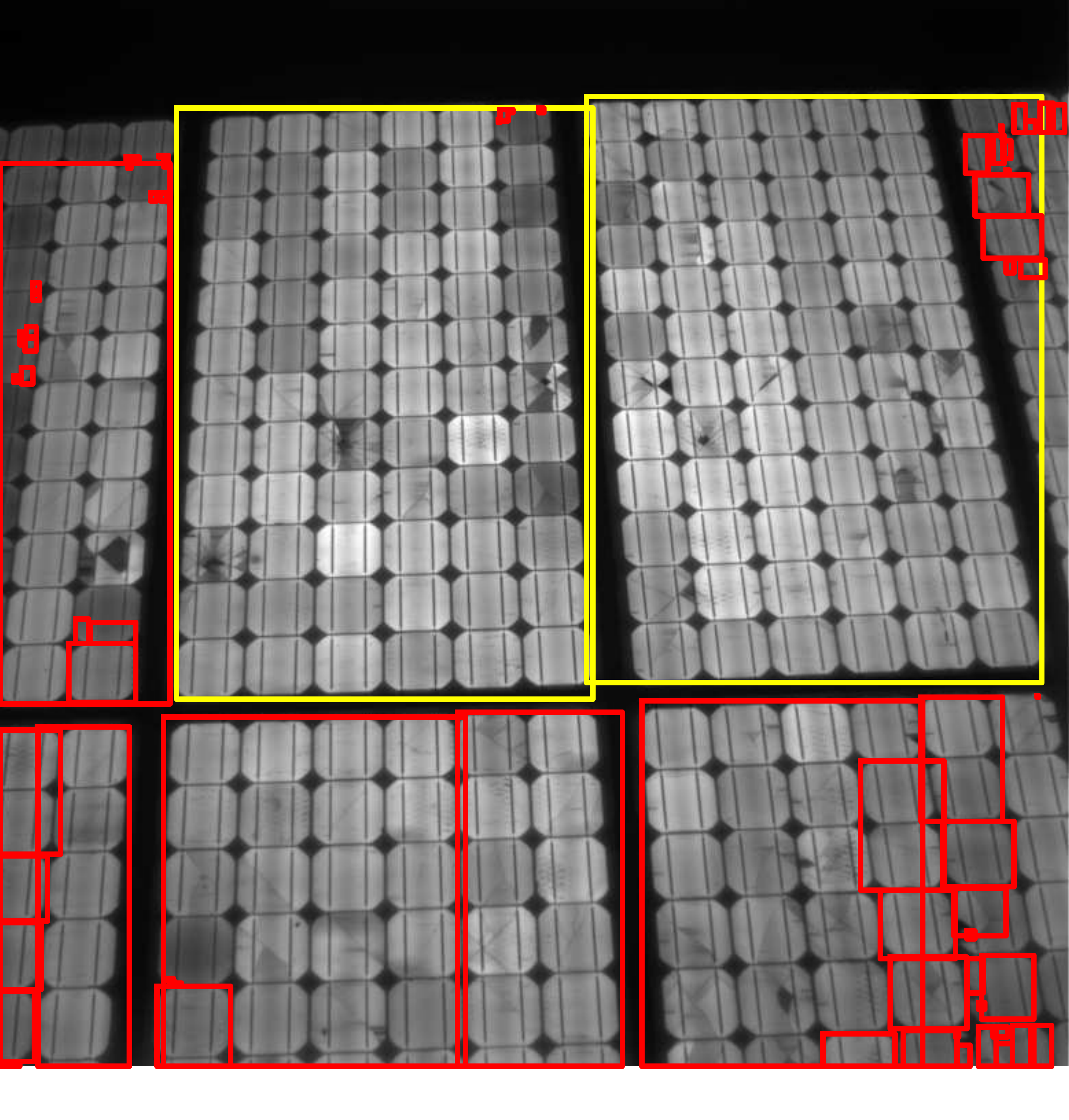}
    \end{subfigure} \\%
    \begin{subfigure}{.24\textwidth}
        \includegraphics[width=\linewidth, angle=270]{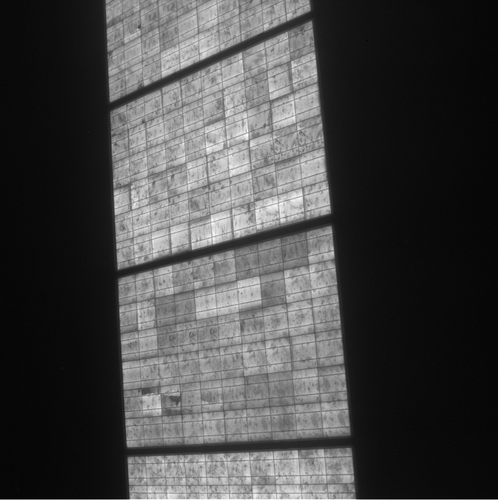}
        \caption{}
        \label{fig:preproc-original}
    \end{subfigure}%
    \hfill%
    \begin{subfigure}{.24\textwidth}
        \includegraphics[width=\linewidth, angle=270]{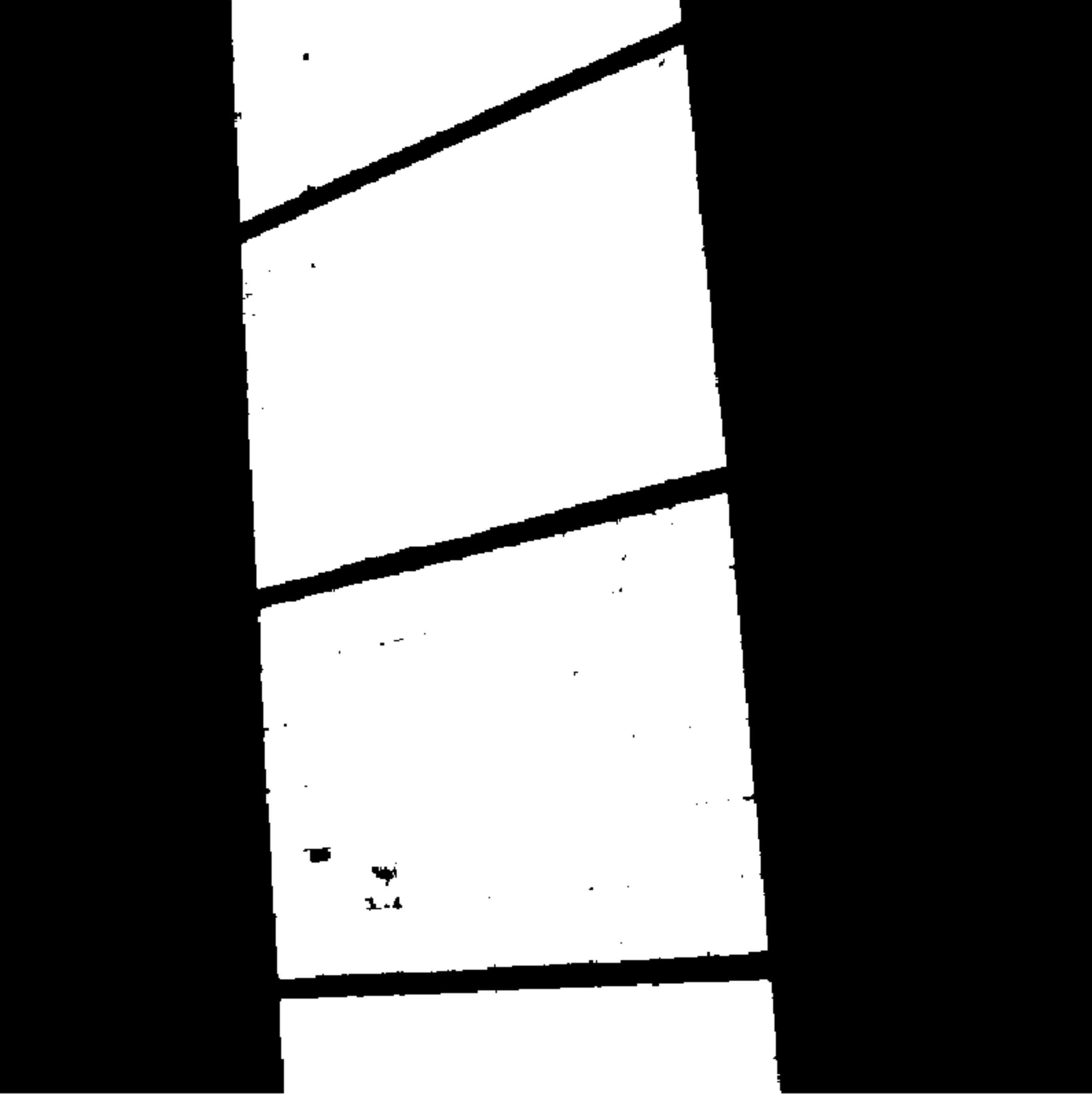}
        \caption{}
        \label{fig:preproc-binary}
    \end{subfigure}%
    \hfill%
    \begin{subfigure}{.24\textwidth}
        \includegraphics[width=\linewidth, angle=270]{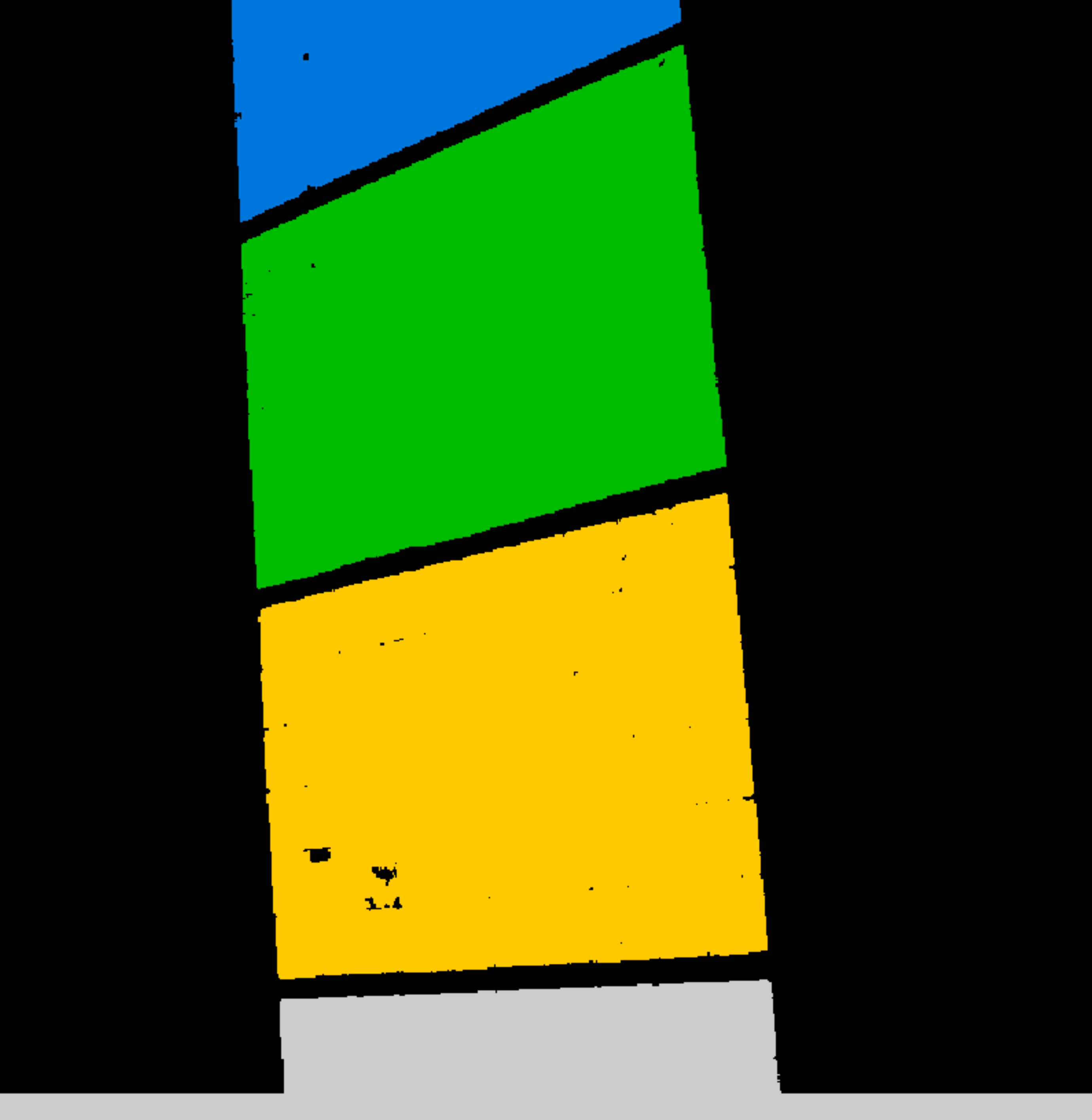}
        \caption{}
        \label{fig:preproc-labeled}
    \end{subfigure}%
    \hfill%
    \begin{subfigure}{.24\textwidth}
        \includegraphics[width=\linewidth, angle=270]{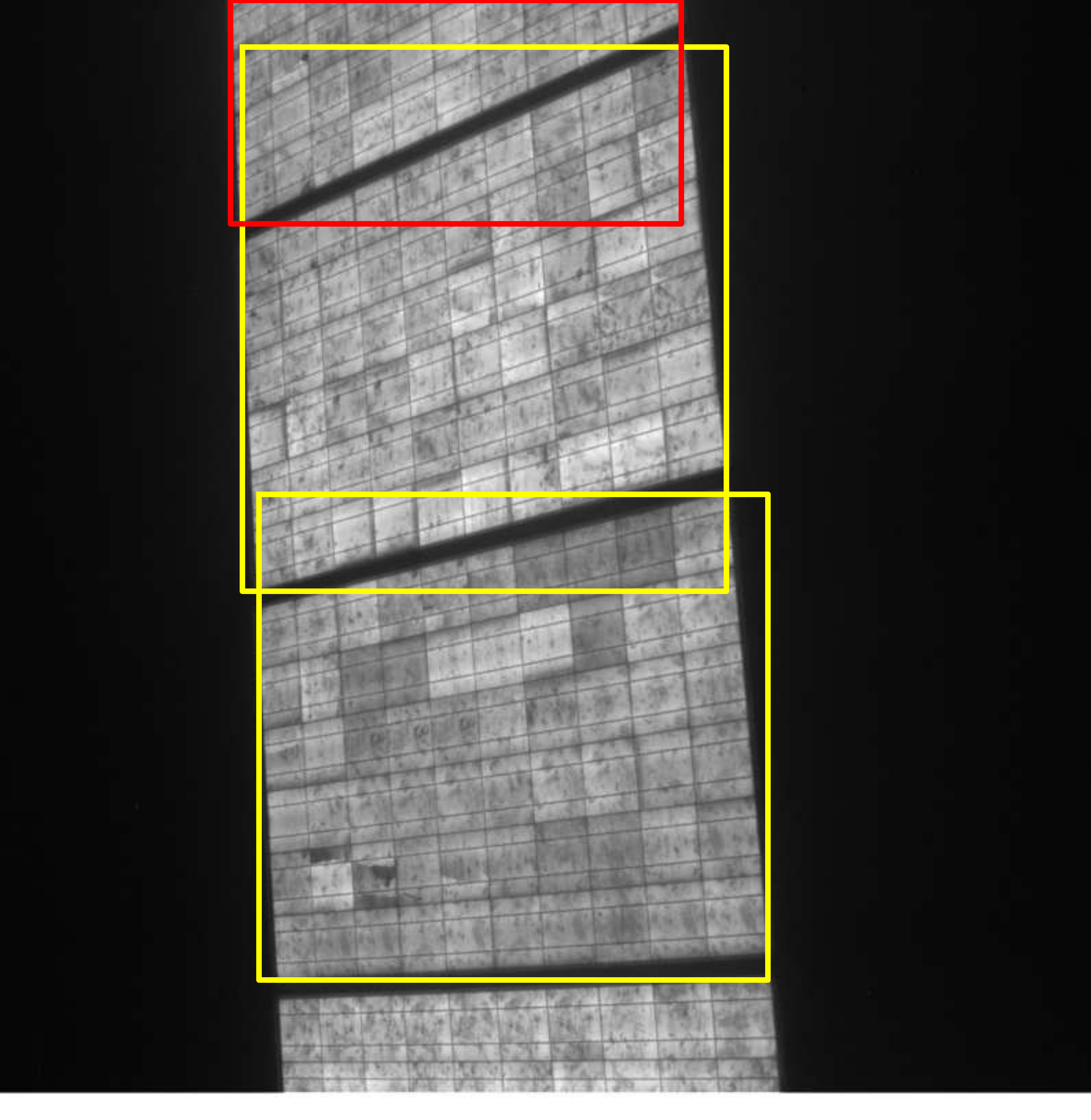}
        \caption{}
        \label{fig:preproc-boxes}
    \end{subfigure}%
    \caption{Preprocessing of multi-module measurements (top: module type \typee, bottom: module type \typeh): Original image (\cref{fig:preproc-original}) is first downsampled and binarized (\cref{fig:preproc-binary}). Next, connected component analysis is applied (\cref{fig:preproc-labeled}). Finally, bounding boxes of connected components are computed and implausible detections are rejected (\cref{fig:preproc-boxes}). Here, rejected detections are depicted by a red bounding box, whereas accepted detections are shown yellow.}
    \label{fig:preproc}
\end{figure*}

In this work, we use measurements from multiple sources. These include measurements taken under controlled lab conditions, as well as measurements taken on-site. For on-site measurements, it is common practice to capture multiple modules in a single image in order to reduce the overall number of measurements~(\cf~\cref{fig:data-examples}). In order to assess the power of a single module using these measurements, a localization of module instances is required. We propose a fast and straight-forward preprocessing pipeline to locate \ac{PV} modules in \ac{EL} measurements.

\subsection{Preprocessing}\label{subsec:preprocessing}

During preprocessing, we make use of the fact that the background in \ac{EL} measurements has a weak intensity, whereas the modules appear as densely connected regions with a high intensity. In addition, modules are usually clearly separated by a distinguished margin. Therefore, background and modules are easily separable by thresholding. We propose a five-step approach:
\begin{enumerate}
    \item\label{itm:downscale} The original measurement is downscaled by~\prescale\deletion[ref=co040,label=ch033]{in order to speed up subsequent computations and decrease the probability of modules to become disconnected by small cracks or dirt}.
    \item Binarization is performed using Otsu's method~\cite{otsu1979threshold}. As a result, the modules are clearly separated~(\cf~\cref{fig:preproc-binary}).
    \item Region candidates are computed by connected component labelling~(\cf \cref{fig:preproc-labeled}) and subsequent region proposal~(\cf~\cref{fig:preproc-boxes})~\cite{jahne2013digitale}.
    \item Final module regions are obtained by rejecting implausible regions that do not adhere to simple constraints~(\cf~\cref{fig:preproc-boxes,subsec:preproc-region-constraints}) and modules are segmented.
    \item \change[ref=co009,label=ch004]{Modules are rectified}{Perspective distortion of module images is removed} using the method proposed by \etal{Hoffmann}~\cite{hoffmann2019fast}.
\end{enumerate}
\addition[ref=co040,label=ch032]{The downscaling in~\cref{itm:downscale} is necessary, since the method is based on binarization and blob detection: Because we only aim at detecting entire modules, small structures such as cracks, busbars, cell borders or dirt are not of any interest to the detection at all. However, they regularily cause modules to appear as multiple seperated blobs in the binarized images, which prevents the accurate detection of the modules. Fortunately, downscaling the images turned out as an efficient and simple method to counter these effects. At the same time, it speeds up subsequent computations.}

\subsection{Region constraints}\label{subsec:preproc-region-constraints}

The proposed pipeline results in a large number of false positives. This is due to noise in the background, parts of a module that are disconnected or modules that are only partially visible. The following constraints are applied to reject false positives:
\begin{enumerate}
    \item Modules that are not completely visible in the measurement are not of interest in this work. Therefore, any detections that touch the boundary of a measurement are rejected.
    \item Most of the outliers have a very small area, compared to the area of inliers. Therefore, any detections with an area small than $\pretsize \cdot a_{\text{max}}$ are rejected. Here, $a_{\text{max}}$ refers to the area of the largest detection in a measurement and \pretsize is a hyperparameter.
\end{enumerate}

As a result, we obtain a method that has only two hyperparameters: The scale \prescale, at which measurements are processed and the minimum area relative to the maximum area \pretsize.

\subsection{Experiments}\label{subsec:preproc-experiments}

The hyperparameters of this method require proper tuning. To this end, we set up a separate dataset using a total of \num{37} measurements from the original on-site measurements. In summary, these measurements show two to seven completely visible modules each. We manually annotate bounding boxes for every completely visible module. Finally, we split the dataset into $25$ training measurements with $75$ completely visible modules and $12$ test measurements with $31$ modules.

For hyperparameter tuning, we use the \optuna with default settings. Since we are mostly interested in a high detection rate and low false positives, we resort to maximizing the $F_1$ score. \addition[ref=co029,label=ch016]{Here, the $F_1$ score is defined as the harmonic mean of precision and recall. Hence, it jointly accounts for the number of false positive and false negative detections.} In order to compute the $F_1$ score for a set of detected and ground truth object boxes, it is necessary to assess, if a ground truth box has been detected and, vice versa, if a detection corresponds to a ground truth box or is a false positive. To this end, we state that a module is detected, if the \ac{IoU} between detected and ground truth bounding box is at least $\pretiou=0.9$. \addition[label=ch005,ref=co010]{Here, the \ac{IoU} is defined as the size of the intersection area divided by the size of the union area between two bounding boxes. Hence, the \ac{IoU} is maximized, if two bounding boxes overlap perfectly.} Then, we calculate the $F_1$ score and perform the optimization on the training set. Overall, the problem is optimized using 30 trials with the default tree-structured Parzen window trial generator. We find that $\prescale=0.23$ and $\pretsize=0.42$ gives the best results on our training split with an $F_1$ score of \num{1}.

\subsection{Results}

The detection performance on the test split evaluates as $F_1 = 0.94$ using the same threshold $\pretiou=0.9$. Detailed results on the detection performance using different thresholds can be found in~\change[]{\ref{app:detection-results}}{\cref{app:detection-results}}. We find that the method performs very well for $\pretiou < 0.9$ and performs slightly worse for larger values of \pretiou. However, given that we are mostly interested in a high detection rate rather than very accurate detections, the performance is very good. Further, the method only takes about \SI{322}{\milli\second} (i7-8650U CPU) for a single image. \addition[ref=co007,label=ch041]{Please also note that this method does not make any assumptions on the number of modules visible in a single image. Hence, \prescale is estimated independent of the number of modules, although the optimal scale is dependent on the number of modules or the resolution of a single module.}

\section{Power prediction}\label{sec:powerprediction}

We estimate the power \pmpp at the maximum power point (STC conditions) of a module using a single \ac{EL} image and the nominal power according to the data sheet \pnom of the latter. \change[ref=co002,label=ch002]{To the best of our knowledge, there is not prior work on this task. Therefore, we}{We} set up a baseline using \ac{SVR} in~\cref{subsec:pp-baseline}. We describe our \ac{DL}-based approach in~\cref{subsec:pp-dl}. In order to make the results using \ac{SVR} and \ac{DL} comparable, we stick to a common experimental procedure, which we describe in~\cref{subsec:pp-protocol}. Finally, in~\cref{subsec:ex-dl-results}, we compare both approaches and report experimental results.

Throughout this work, we assume that \addition[ref=co043,label=ch034]{the nominal power} \pnom is known for every module. Then, we estimate the power relative to \pnom, \ie
\begin{equation}\label{eq:pmpp}
    \pmpp = \prel \cdot \pnom
\end{equation}
and denote estimates of \addition[ref=co046,label=ch037]{the relative power} \prel by \prelest. Note that, for this dataset, \prel roughly correlates to the amount of the inactive area of a module \addition[ref=co043,label=ch035]{($r=0.90$)}. \addition[ref=co044,label=ch036]{While we cannot say, if the slope of a linear model predicting \prel from the amount of inactive area is the same for every module type, we expect that the observed correlation between the two quantities holds for other module types as well. This is also supported by the work of \etal{Schneller} that comes to a similar observation using different data~\cite{schneller2018electroluminescence}.}

\subsection{Evaluation protocol}\label{subsec:pp-protocol}

The overall goal is to find a model that predicts \prel with a low average error and a small number of outliers. For evaluation purposes, we report the \ac{MAE} over all samples, as well as the \ac{RMSE}. The \ac{MAE} is given by
\begin{equation}\label{eq:mae}
    \mae = \frac{1}{\nsamples} \sum_{i=1}^\nsamples |\prel^{(i)}-\prelest^{(i)}|\,
\end{equation}
while the \ac{RMSE} is computed as
\begin{equation}\label{eq:rmse}
    \rmse = \frac{1}{\nsamples} \sqrt{ \sum_{i=1}^\nsamples \left(\prel^{(i)}-\prelest^{(i)}\right)^2 }\,.
\end{equation}

Since the dataset is relatively small, making statistically significant statements on model performance using a conventional train/test split of the dataset is hard, since results are necessarily computed on a relatively small test set. To overcome this limitation, we conduct a \ac{CV}, such that the complete dataset is used for testing. We initially split the dataset into 5 folds using stratified sampling, such that the overall distribution of relative powers is preserved between folds. Since a stratified sample requires distinct class labels, we discretize \prel in \num{20} distinct bins, where each bin has a range of \SI{5}{\percent}. Further, we make sure that none of the solar module instances ends up in two different folds.

The baseline method as well as the \ac{DL}-based methods have hyperparameters that need to be tuned properly. In order to enable a fair comparison between methods, we establish a standard protocol. We perform the hyperparameter optimization using \optuna with the default settings and \num{250} iterations in every fold of the \ac{CV} to make sure that no test data is used for hyperparameter optimization. Then, we split the training folds into a training and a validation set. While the first one is used to determine model parameters, the latter one is used to optimize hyperparameters. We empirically found that the validation set needs to be large enough to obtain stable results. To this end, we use \SI{40}{\percent} of the training data for validation. Again, we perform a stratified split, such that the label-distributions are similar in training and validation set.

\subsection{Baseline}\label{subsec:pp-baseline}

In this section, we propose a baseline approach to estimate the \pmpp from a single \ac{EL} image. We train the \ac{SVR} using features extracted from the \ac{EL} measurements.

\subsubsection{Feature extraction}

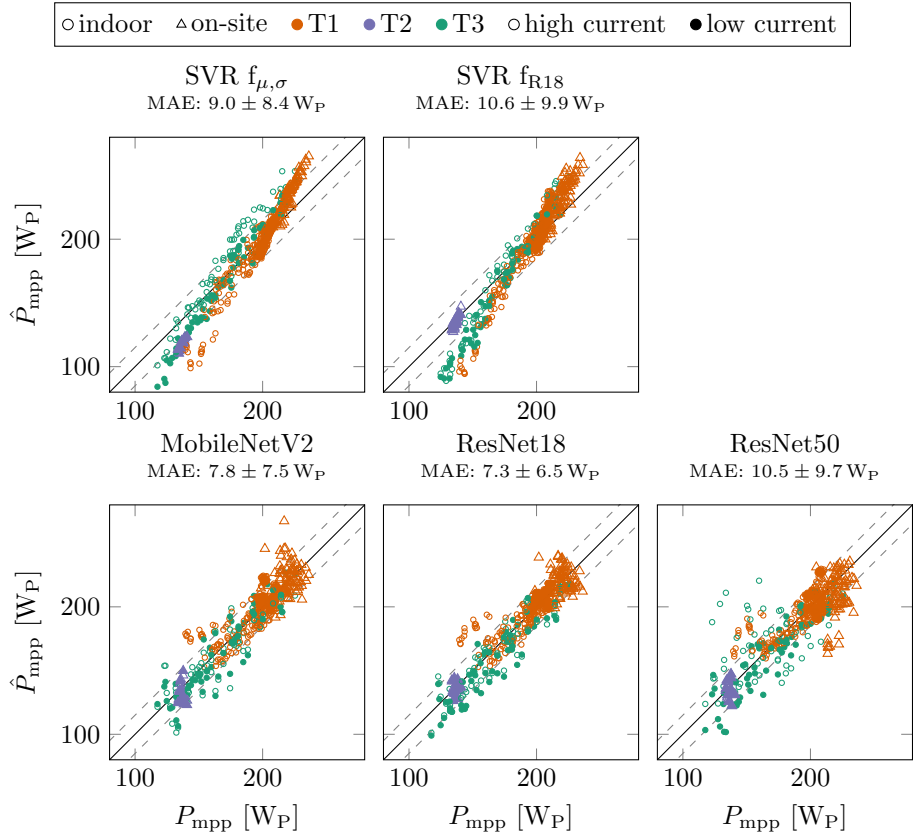
\begin{figure*}[tp]
    \pgfplotsset{
        compat=newest,
        every axis/.append style={
            scatterplotnew style,
            xtick={100,200},
            ytick={100,200},
            title style={align=center, font=\linespread{0.8}\selectfont},
        }
    }

    \centering
    \begin{NoHyper}\tikzexternaldisable\ref{scatter-all-legend}\tikzexternalenable\end{NoHyper}
    \vspace{.1cm} \\

    \tikzsetnextfilename{scatter_all}%
    \begin{tikzpicture}
        \begin{groupplot}[
            group style={
                group size=3 by 2,
                vertical sep=1.5cm,
                ylabels at=edge left,
                xlabels at=edge bottom,
                yticklabels at=edge left,
                horizontal sep=.25cm,
            },
            width=1/2.3*\textwidth-8.5pt,
            height=1/2.3*\textwidth-8.5pt,
        ]

            \nextgroupplot[
                title={\ac{SVR}~\fmstd\\{\scriptsize\acs{MAE}:~\resultsmaew{svrmstd}}},
            ]
                \addplot [
                    scatter samples,
                ]table [x=peak_power, y=predicted_power, meta=source_key, col sep=comma]{csv_results/038_svr_cv/cv_mean_std_test_errors_all.csv};
                \scattercommon
            
            \nextgroupplot[
                title={\ac{SVR}~\fresnet\\{\scriptsize\acs{MAE}:~\resultsmaew{svrpt}}},
                legend columns=-1,
                legend to name=scatter-all-legend,
                legend entries={indoor,on-site,\typet,\typee,\typeh,high current,low current},
                legend style={/tikz/every even column/.append style={column sep=0.25cm}},
            ]

                \addlegendimage{only marks,mark=o,black}; \label{pgfplots:mark-indoor}
                \addlegendimage{only marks,mark=triangle,black}; \label{pgfplots:mark-outdoor}
                \addlegendimage{only marks,mark=*,colort1}; \label{pgfplots:mark-typet}
                \addlegendimage{only marks,mark=*,colore3}; \label{pgfplots:mark-typee}
                \addlegendimage{only marks,mark=*,colorh2}; \label{pgfplots:mark-typeh}
                \addlegendimage{only marks,mark=o,black}; \label{pgfplots:mark-highcurrent}
                \addlegendimage{only marks,mark=*,black}; \label{pgfplots:mark-lowcurrent}
                \addlegendimage{no marks,black}; \label{pgfplots:line-ideal}
                \addlegendimage{no marks,dashed,white!50!black}; \label{pgfplots:line-dashed}

                \addplot [
                    scatter samples,
                ]table [x=peak_power, y=predicted_power, meta=source_key, col sep=comma]{csv_results/038_svr_cv/cv_pretrained_test_errors_all.csv};
                \scattercommon

            \nextgroupplot[group/empty plot]

            \nextgroupplot[
                title={\mobilenet\\{\scriptsize\acs{MAE}:~\resultsmaew{mobilenet}}},
            ]
            \addplot [
                scatter samples,
            ]table [x=peak_power, y=predicted_power, meta=source_key, col sep=comma]{csv_results/037_cv_results/mobilenet_v2_all.csv};
            \scattercommon

            \nextgroupplot[
                title={\resneta\\{\scriptsize\acs{MAE}:~\resultsmaew{resnet18}}},
            ]
            \addplot [
                scatter samples,
            ]table [x=peak_power, y=predicted_power, meta=source_key, col sep=comma]{csv_results/037_cv_results/resnet18_all.csv};
            \scattercommon
            
            \nextgroupplot[
                title={\resnetb\\{\scriptsize\acs{MAE}:~\resultsmaew{resnet50}}},
            ]
            \addplot [
                scatter samples,
            ]table [x=peak_power, y=predicted_power, meta=source_key, col sep=comma]{csv_results/037_cv_results/resnet50_all.csv};
            \scattercommon

        \end{groupplot}
    \end{tikzpicture}

    \tikzexternaldisable
    \caption{Distribution of estimation errors of all methods. We aggregate the results from all folds of the \acf{CV}, such that all samples of the dataset are shown here. For the samples, we distinguish between indoor~\ref{pgfplots:mark-indoor} and on-site \ref{pgfplots:mark-outdoor} measurements by marker shape. Further, we distinguish between module type \typet~\ref{pgfplots:mark-typet}, \typee~\ref{pgfplots:mark-typee} and \typeh~\ref{pgfplots:mark-typeh} by marker color. Finally, we differentiate measurements taken at high current~\ref{pgfplots:mark-highcurrent} or low current~\ref{pgfplots:mark-lowcurrent} by marker filledness. For better visualization, the ideal regression line is shown~\ref{pgfplots:line-ideal}. Furthermore, \ref{pgfplots:line-dashed} indicates the \SI[separate-uncertainty=true, multi-part-units=single]{0(15)}{\wattpeak} isoline.}
    \tikzexternalenable
    \label{fig:scatter-all}
\end{figure*}

\addition[label=ch007,ref=co012]{To apply power estimation using the \ac{SVR}, we map the data to a low dimensional feature representation. This representation is defined such that as much as possible information that is required to estimate the power, is preserved. }\change[ref=co026,label=ch044]{In this work, we compare two types of features. The }{In previous works, Zernike moments and Wavelet features have been successfully applied to recognize textured defects such as cracks on \ac{PV} modules~\cite{li2012wavelet,farress2017defect}. However, we know that the power of a module is largely dominated by fractures, since our dataset is carefully compiled to adhere to this property. Since fractures are mainly blob-like defects, we use the }measurement mean and standard deviation\addition[ref=co026]{, which is a good representation for that}\deletion[ref=co026]{ are plausible features, since \prel roughly correlates to the amount of inactive area, which is well reflected by these features}. Furthermore, it has been shown that features extracted from \change[label=ch009,ref=co013]{pretrained \ac{DL} models are useful in other domains than the training domain as well}{models pretrained on the ImageNet~\cite{deng2009imagenet} dataset are useful for machine learning tasks on different data as well}~\cite{oquab2014learning,sharif2014cnn}. Therefore, we include features extracted from a pretrained \resneta as well~\cite{he2016identity}. \addition[ref=co029,label=ch017]{The ResNet architecture has become popular for many computer vision problems. It is available in different configurations featuring between $18$ and $152$ layers. It was the first architecture that introduced residual connections between subsequent layers, which mainly solved the issue of vanishing gradients. As a result, it enables to increase the number of layers of a \ac{DL} architecture.} To compute \addition[ref=co029,label=ch018]{the \resneta-features} \fresnet, we convert measurements into RGB images and perform channel-wise normalization using the per-channel statistics. Hence, we make sure that the statistics of the measurements match those of the ImageNet~\cite{deng2009imagenet} dataset that has been used during training. For the pretrained \resneta, we use the model available in \addition[ref=co029,label=ch019]{the open-source \ac{DL} framework} \pytorch and drop the fully connected layer to obtain features directly after global average pooling. On the other hand, \addition[ref=co029,label=ch020]{the features composed of the mean and standard deviation of intensities} \fmstd is computed without further preprocessing.

The feature extraction is followed by a \change[label=ch010,ref=co015]{standardization}{normalization}. Here, we compute the mean and standard deviation of every feature independently \change[label=ch008,ref=co012]{for the features extracted from the training data}{using all samples from the training data}. We use these statistics to normalize training, as well as test features.

\subsubsection{Experimental procedure}

The \ac{SVR} regression aims to minimize the absolute regression error
\begin{equation}\label{eq:svr-objective}
    |\perrrel| = |\prel-\prelest|
\end{equation} 
for those samples that have $\perrrel > \epsilon$, while maintaining a regression model with a small Lipschitz constant. Here, $\epsilon$ is the width of the epsilon-tube within which errors do not contribute to the loss of the \ac{SVR} objective function~\cite{drucker1997support}. We use the implementation from the \sklearn with the default \change[ref=co029,label=ch021]{RBF}{radial basis function} kernel. For the width of the epsilon-tube and the regularization constant, we conduct a hyperparameter optimization in every fold as described in~\cref{subsec:pp-protocol}. 

\subsection{Deep-learning}\label{subsec:pp-dl}

In this section, we introduce the \ac{DL}-based approach to predict \pmpp given an \ac{EL}-measurement as well as the nominal power \pnom of the module. We present a straightforward approach and use standard \ac{DL}-architectures that have been trained on ImageNet~\cite{deng2009imagenet} to perform regression of \prelest (see~\cref{eq:pmpp}). We detail our methodology in~\cref{subsec:pp-dl-pipeline} and focus on the pipeline that empirically worked best. In~\cref{subsec:ex-dl-procedure} we explain, how we tuned the hyperparameters for different \ac{DL}-models.

\subsubsection{Method}\label{subsec:pp-dl-pipeline}

As \acp{DNN} are usually trained end-to-end, meaning that the feature extraction is part of the training process, there are only a few design choices that need to be made. These include
\begin{enumerate*}
    \item the preprocessing that is applied to the raw measurements
    \item the \ac{DL}-models that are used
    \item the loss function
    \item the optimization method
\end{enumerate*}. We detail our choices in the next paragraphs:

\paragraph{Preprocessing}

\begin{figure*}[t]
    \pgfplotsset{
        compat=newest,
        every axis/.append style={
            width=5.0cm,
            height=4.5cm,
            scale only axis,
        }
    }
    \centering%
    \subcaptionbox{Sensitivity of the error with respect to variations in measurement scale.}{
        \tikzsetnextfilename{dl_scale}%
        \begin{tikzpicture}[object/.style={thin,double,<->}]
            \begin{axis}[
                xlabel={scale [\si{\px}]},
                ylabel={\ac{MAE}~[\si{\percent}]},
                axis y discontinuity=crunch,
            ]
                \addplot+[
                    error bars/.cd,
                    y dir=both,
                    y explicit,
                ] table [x=img_size, y expr=100*\thisrow{mean_mae}, y error expr=100*\thisrow{std_mae}, col sep=comma]{csv_results/004_check_best_scale/error_by_scale.csv};
            \end{axis}
        \end{tikzpicture}%
        \label{fig:ex-dl-scale}%
    }%
    \hfill%
    \subcaptionbox{Sensitivity of the error with respect to variations in the normalization scheme.}{%
        \tikzsetnextfilename{dl_normalization}%
        \begin{tikzpicture}
            \begin{axis}[
                xlabel={epochs},
                axis y discontinuity=crunch,
            ]
                \addplot [
                    only marks,
                    scatter/classes={
                        globalstandardization={mark=square*,blue},
                        imagewisestandardization={mark=triangle*,red},
                        imagewisezcawhitening={mark=o,black}
                    },
                    scatter,
                    scatter src=explicit symbolic,
                    axis y discontinuity=crunch,
                    error bars/.cd,
                    y dir=both,
                    y explicit,
                    x dir=both,
                    x explicit,
                ] table [y expr=100*\thisrow{mae_mean}, x=n_epochs_mean, y error expr=100*\thisrow{mae_std}, x error=n_epochs_std, meta=normalize, col sep=comma]{csv_results/012_compare_normalization/results.csv};
                \legend{global std.,measurement std.,\acs{ZCA} whitening};
            \end{axis}
        \end{tikzpicture}%
        \label{fig:ex-dl-normalization}%
    }%
    \caption{We conduct a series of initial experiments to determine the best measurement scale and normalization scheme. All experiments were performed using \resneta without further hyperparameter tuning. Every model is trained \num{5} times until convergence. We report the minimum \ac{MAE} on the validation set. Error bars denote the standard deviation of the \ac{MAE} between training runs. In~\cref{fig:ex-dl-scale}, we scale the smallest side of the measurement to the specified value. In~\cref{fig:ex-dl-normalization}, we compare the common global normalization, where all measurements are normalized by the same statistics computed over the whole training set to a measurement-wise approach, where every measurement is normalized using it's own statistics. Finally, we include normalization by patch-wise \ac{ZCA}-whitening. It turns out that a scale of \SI{800}{\px} and the common global normalization give the best results.}
    \label{fig:ex-dl-initial}
\end{figure*}
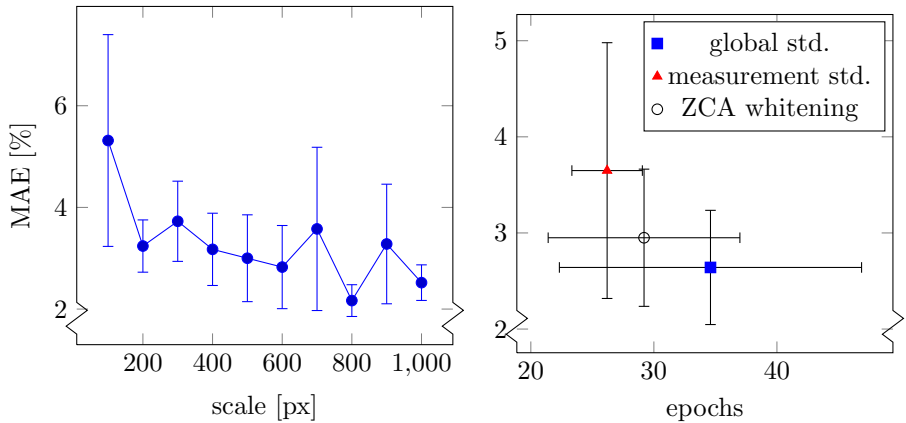%

Although we only use fully-convolutional networks, it is common to limit the resolution of input measurements to reduce the computational effort. We rescale measurements such that the smallest side equals \SI{800}{\px}. Then, we normalize the measurements using the common normalization with global statistics. To this end, we compute the mean \mymean and standard deviation \mystd of photon counts using our data. A sample \mysample is then normalized according to
\begin{equation}\label{eq:normalization}
    \mysample' = \frac{\mysample-\mymean}{\mystd}\,.
\end{equation}
The choices of scale and normalization scheme are verified by initial experiments reported in~\cref{fig:ex-dl-initial}.

\addition[ref=co028,label=ch015]{Furthermore, we apply online data augmentation during training. As already reported by \etal{Deitsch}~\cite{deitsch2019automatic} for the case of single cells, light data augmentation is sufficient, since the segmented modules are mostly fixed in orientation and position due to the module detection and removal of perspective distortion. We apply random horizontal and vertical flips, random rotation by at most $\pm 10^\circ$, random translation by $5\%$ of the image size at maximum and random downscaling to $80\%$ of the image size at minimum. The augmented images are zero-padded to match the original image size.}

\paragraph{\ac{DL} models}

In many recent works, it has been shown that applying transfer learning is advantageous over training models from scratch, even if source and target domain differ~\cite{yosinski2014transferable}. Therefore, we resort to using standard architectures, where pretrained weights are readily available. In this work, we focus on three different architectures: First, we include a small and a larger network from the ResNet architecture family (\resneta and \resnetb)~\cite{he2016identity}. This allows us to investigate, if deeper architectures are beneficial for the task at hand. In addition, we include the \mobilenet architecture~\cite{sandler2018mobilenetv2}, which is specifically designed for high throughputs at inference time, since this is a major benefit for practical applications.

For every one of those models, \addition[ref=co019,label=ch022]{the final layer is composed by a matrix multiplication that maps the high dimensional feature representation to the correct output size. This layer is often referred to as the \ac{FC} layer.} We replace the final \ac{FC} layer by a randomly initialized \ac{FC} layer with a single output. We do not perform any non-linear activation on the output. Hence, this corresponds to a linear regression of \prel using the features from previous layers.

\paragraph{Loss function}

Common loss functions for regression problems include the \ac{MSE} and \ac{MAE}. By definition, \ac{MSE} puts a higher weight on outliers, resulting in a model that performs better for underrepresented cases and worse for overrepresented cases. Since we are interested in a model that performs well over a large range of samples, we minimize the \ac{MSE}.

\paragraph{Optimization}

A huge variety of optimization methods has been proposed in the literature. For this work, we decided to use some of the most prevalent approaches: We use \ac{SGD} with momentum $\mymomentum = 0.9$ and weight decay \mywd and properly tune the learning rate \mylr, batch size \mybs and weight decay \mywd independently for every architecture. For the learning rate schedule, we reduce the learning rate as soon as the validation loss does not decrease for \num{20} epochs. \addition[label=ch011,ref=co019]{Here, an epoch refers to training once on the complete training dataset.}

\subsubsection{Experimental procedure}\label{subsec:ex-dl-procedure}

Proper tuning of hyperparameters is crucial in order to perform a fair comparison of different architectures. We use the \optuna to determine the optimal values for \mylr, \mybs and \mywd individually for every architecture and fold of the \ac{CV}, as described in~\cref{subsec:pp-protocol}. The results are summarized by~\cref{tab:dl-hyperparams}. We find that optimal parameters are relatively similar across architectures.

As soon as the optimal hyperparameters are determined, we use them to train a final model for every fold that is then tested using the test data of the respective fold. As opposed to the hyperparameter search, we now use~\SI{80}{\percent} of the data for training and only~\SI{20}{\percent} for validation. We further apply early stopping after \num{40} epochs without improvement on the validation set and use the checkpoint that performed best on the validation set for testing.

\subsection{Results}\label{subsec:ex-dl-results}

\begin{figure*}[t]
    \centering
    \tikzsetnextfilename{boxplots_all}%
    \begin{tikzpicture}%
        \pgfplotsset{
            every axis plot/.append style={mark options={scale=0.5}}
        }%
        \begin{axis}[
            boxplot = {
                draw direction=y,
                box extend=0.5,
            },
            width=\textwidth,
            height=7cm,
            minor y tick num=5,
            ymajorgrids=true,
            major grid style={color=black!50!white},
            minor grid style={color=black!10!white},
            ylabel={$\lvert\perrrel\rvert$~[\si{\percent}]},
            ymax=15,
            xtick={1,2,...,5},
            xticklabels={%
                {SVR \fmstd},%
                {SVR \fresnet},%
                {\mobilenet},%
                {\resneta},%
                {\resnetb},%
            },
            x tick label style={
                text width=2.5cm,
                align=center,
            },
            cycle list={
                {draw=black,fill=Paired-A},
                {draw=black,fill=Paired-A}
            },
        ]%
            \addlegendimage{mark=*,fill=white}

            \foreach \model in {svrmstd,svrpt,mobilenet,resnet18,resnet50} {%
                \addplot+ [boxplot/data={\thisrow{error}}, discard boxplot if not={model}{\model}]
                    table[col sep=comma]{csv_results/039_summary/boxplots.csv};%
            }%
            \legend{outlier};%
        \end{axis}%
    \end{tikzpicture}%
    \caption{Distributions of sample errors over all 5 folds of the \ac{CV}. The errors are computed on the testset of every fold. Note that the $y$ axis has been cut at $\lvert\perrrel\rvert = 15\%$ for better visualization.}
    \label{fig:boxplots-mae-rmse-all}
\end{figure*}
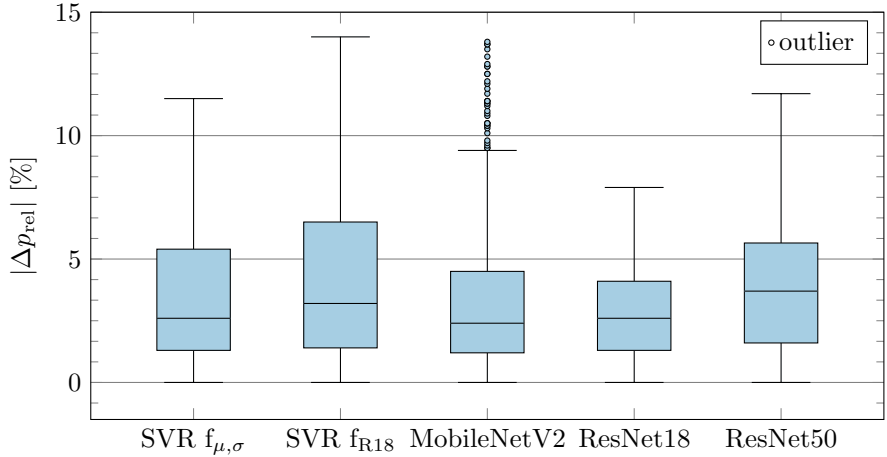

\begin{table}[t]
    \centering
    \csvloop{
        file=csv_results/039_summary/params_summary.csv,
        head to column names,
        before reading=\sisetup{table-format=1.2e-1,separate-uncertainty=true,table-figures-uncertainty=3,table-align-uncertainty=true, scientific-notation=true},
        tabular={lS[table-format=1]SS[table-figures-uncertainty=4]},
        table head=\toprule & \textbf{\mybs} & \textbf{\mywd} & \textbf{\mylr} \\ \midrule,
        command=\texmodel & \batchsizemedian & \weightdecaymean(\weightdecayuncertainty)\weightdecayexp & \learningratemean(\learningrateuncertainty)\learningrateexp,
        table foot=\bottomrule
    }
    \caption{Results from the hyperparameter optimization using \optuna. The hyperparameter optimization has been performed using $250$ iterations per model. We optimize the the batch size sampling from $\mybs \in \left[8,\,16,\,32,\,64\right]$, the learning rate sampling from $\mylr \in \left[\num{1e-5},\,\num{1e-1}\right]$ and the weight decay sampling from $\mywd \in \left[\num{1e-3},\,\num{1e-1}\right]$ in logspace. Sampling is performed using the default tree-structured Parzen window approach and unsuccessful trials are pruned.}
    \label{tab:dl-hyperparams}
\end{table}

\begin{table}[t]
    \centering
    \csvloop{
        file=csv_results/039_summary/summary.csv,
        head to column names=false,
        column names={model=\csvmodel,maeW=\csvmaeW,mae=\csvmae,stdW=\csvstdW,std=\csvstd,rmseW=\csvrmseW,rmse=\csvrmse,texmodel=\csvtexmodel,uncertainty=\csvuncertainty,uncertaintyW=\csvuncertaintyW},
        before reading=\sisetup{table-format=2.1,round-mode=places,round-precision=1,table-number-alignment=left},
        tabular={lS[table-format=1.1,separate-uncertainty=true,table-figures-uncertainty=3,table-align-uncertainty=true]S[table-format=1.1,table-number-alignment=left]S[separate-uncertainty=true,table-figures-uncertainty=3,table-align-uncertainty=true]S[table-format=1.1,table-number-alignment=left]},
        table head={\toprule \linespread{0.1} & \text{MAE}~[\si{\percent}] & \text{RMSE}~[\si{\percent}] & \text{MAE}~[\si{\wattpeak}] & \text{RMSE}~[\si{\wattpeak}] \\ \midrule},
        command=\csvtexmodel & \csvmae(\csvuncertainty) & \csvrmse & \csvmaeW(\csvuncertaintyW) & \csvrmse,
        table foot=\bottomrule
    }
    \caption{Quantitative results computed by \ac{CV}. We show the \ac{MAE} and the \ac{RMSE} averaged over the test sets of every fold of the \ac{CV}. In addition, the standard deviation of errors is shown. Finally, we explicitly show the relative error, as used in training as well as the absolute error in [\si{\wattpeak}] for better interpretability.}
    \label{tab:dl-results}
\end{table}

\begin{table}[tp]
    \newcommand{\outdoor}{on-site}
    \newcommand{\indoor}{indoor}
    \newcommand{\high}{high current}
    \newcommand{\low}{low current}
    \centering
    \csvloop{
        file=csv_results/039_summary/summary_by_subset.csv,
        head to column names,
        before reading=\sisetup{table-format=2.1,round-mode=places,round-precision=1,table-number-alignment=left},
        tabular={lS[table-format=1.1,separate-uncertainty=true,table-figures-uncertainty=3,table-align-uncertainty=true]S[table-format=1.1,table-number-alignment=left]S[separate-uncertainty=true,table-figures-uncertainty=3,table-align-uncertainty=true]SS[table-format=3]},
        table head={\toprule \linespread{0.1} & \text{MAE}~[\si{\percent}] & \text{RMSE}~[\si{\percent}] & \text{MAE}~[\si{\wattpeak}] & \text{RMSE}~[\si{\wattpeak}] & \nsamples \\ \midrule},
        command={\texsubset & \mae(\uncertainty) & \rmse & \maeW(\uncertaintyW) & \rmseW & \samplecnt},
        table foot=\bottomrule,
    }
    \caption{Results on different subsets of the data using \resneta. We compute the \acf{MAE} and the \acf{RMSE} using the test samples from all 5 folds. In addition, the standard deviation of errors between folds is shown as well as the number of samples \nsamples.}
    \label{tab:subset-results}
\end{table}

In this section, we assess the performance of the architectures for the prediction of \ac{MP} and compare it to the performance of the baseline model. By training \num{5}~\ac{CV} folds with every architecture using the parameters found by hyperparameter optimization, we assess the stability of every method with respect to variations in the training and test data. We report the results on the test folds in~\cref{fig:boxplots-mae-rmse-all} and in~\cref{tab:dl-results}. It turns out that \resneta and \mobilenet have a smaller average \ac{MAE} than the baseline, whereas \resnetb does not perform well. This is explained by the relatively small dataset. Among the \ac{SVR}-based methods, \fmstd has the lowest \ac{MAE}. However, it also shows the strongest variation between folds, indicating that the result is very data dependent. Overall, \resneta gives the lowest \ac{MAE} and \ac{RMSE}.

In terms of variation between \ac{CV} folds, \resnetb turns out most stable. However, the advantage over \resneta is neglectible. We observe that the gap between \ac{MAE} and \ac{RMSE} is smaller for the \ac{DL}-based models compared to the \ac{SVR}-based ones. This is explained by the difference in the loss function, since \ac{SVR} roughly minimizes the \ac{MAE}~(\cf~\cref{eq:svr-objective}), whereas we chose to minimize the \ac{MSE} for the \ac{DL}-based models.

In~\cref{fig:scatter-all}, we show the distribution of errors for all samples in the dataset and all methods. This is obtained by merging all five test folds into a single figure. The comparison shows that the predictions of \resneta are well aligned with the ideal regression line, whereas the predictions using \ac{SVR} are skewed with respect to the ideal line. We conclude that this is caused by the features chosen for the regression model that do not result in a linear regression problem under the RBF kernel. However, initial experiments with other kernels performed even worse. We also compute the $p$-value using a \textit{t}-test to assess, if the \resneta performs significantly better than the other approaches. We find that the performance difference to the \mobilenet is not statistically significant, whereas \resneta performs significantly better than the remaining other methods ($p < 0.0001$).

Finally, in~\cref{tab:subset-results}, we summarize the results for different subsets of the data. It turns out that the prediction is stable across most of the different subsets. The only exception are those modules measured at a low excitation current and modules of type \typeh. Both of them perform worse than the others. Subset~\typeh is composed of modules measured at high and low current. Hence, it contains the same modules twice. We find that modules from subset~\typeh that are measured at high current result in a \ac{MAE} of~\SI{4.3}{\percent}, whereas modules from the same subset measured at a low current result in a \ac{MAE} of~\SI{4.7}{\percent}. Specifically, modules from subset \typeh are mostly underestimated and we find that this effect is more severe for modules measured at a low current. Since those results have been computed using the same module instances and are averaged over all \num{5} folds of the \ac{CV}, we can conclude that the model performs slightly better on measurements taken at a high current, irrespective of the module type. However, this result might be biased by the fact that the dataset is largely dominated by samples measured at a high current.

\subsubsection{Visualization of Class Activation Maps}\label{subsec:ex-dl-cam}

\begin{figure}[t]
    \centering
    \tikzsetnextfilename{cam}
    \begin{tikzpicture}[
    ]
        \begin{axis}[
            width=1.00\textwidth,
            height=0.65\textwidth,
            hide axis,
            y dir=reverse,
            colorbar,
            colormap/viridis,
            point meta min=0,
            point meta max=100,
            colorbar style={
                ytick={0, 25, 50, 75, 100},
                title={[\%]},
            },
            enlargelimits=false,
        ]
            \addplot graphics[xmin=0, ymin=0, xmax=1333, ymax=800] {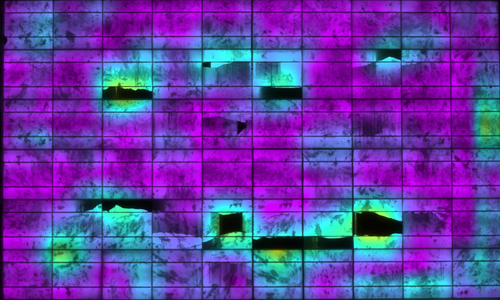};

            \addplot [
                mark=none,
                draw=none,
                nodes near coords={\tiny\num[round-mode=places, round-precision=1, tight-spacing=true]{\pgfplotspointmeta}},
                nodes near coords align={anchor=center},
                nodes near coords style={
                    rectangle,
                    fill=white,
                    fill opacity=0.7,
                    text opacity=1,
                    inner sep=2pt,
                    rounded corners=2pt,
                },
            ] 
            table[
                x=x,
                y=y,
                meta=v,
                col sep=comma,
                point meta=explicit symbolic
                ] {csv_results/042_cam2/180.csv};

        \end{axis}
    \end{tikzpicture}

    \caption{Visualization of \ac{CAM} using a modified \resneta. Note that we color-code the original \ac{EL} measurement with the given colormap using the relative magnitude of $-\fcam$ such that brighter colors correspond to regions with high relative power loss. Since the intensity is given by the original measurement, color appearence does not exactly correspond to the legend, since the legend uses a linear intensity ramp. For every cell, we give the power loss determined by the model in \si{\wattpeak}. This is computed by integrating over the corresponding area of the \ac{CAM}, resulting in the relative power loss, which is converted into the absolute power loss by~\cref{eq:pmpp}.}
    \label{fig:cam}
    
\end{figure}

In previous works, it has been shown that object locations can be extracted from \acp{DNN} trained for object classification~\cite{oquab2015object}. To this end, the activation maps from the last layer that preserves spatial information are used to calculate a heatmap of possible object locations. This concept is now widely known as \acf{CAM}. In a recent work, we used this to segment cracks in \ac{EL} measurements using a \resneta~\cite{mayr2019weakly}.

We propose to build upon this work and use a slightly modified \resneta that allows one to infer the predicted power loss for an arbitrary area by integrating over the respective area of the \ac{CAM}. \addition[ref=co018,label=ch039]{The \resneta is constructed such that it computes \num{512} feature maps given an input image. However, we are interested in a single feature maps that gives the power loss per pixel.} To this end, we append a $1\times1$ convolution using a single kernel to \change[ref=co018]{the feature extraction part of the \resneta, resulting in a single featuremap \fresnetact}{reduce the number of feature maps to a single map \fresnetact}. Since we want every pixel \addition[ref=co018]{of \fresnetact} to quantify the loss of relative power of the corresponding area of the measurement, we need to constrain the resulting map such that it is strictly negative. This is achieved by appending a ReLU, followed by a Sign layer. \addition[ref=co019,label=ch023]{Here, the ReLU refers to a function that is the identity for input values greater zero and zero otherwise, whereas the Sign layer flips the sign of the input.} The resulting feature map \fcam is later used for further assessment. \change[ref=co018]{Finally}{Since \fcam gives the per-pixel power loss}, the total power loss is computed by summing over \fcam. Hence, instead of computing \prelest directly from the activations after \ac{GAP} using a fully connected layer, we now compute \prelest as
\begin{equation}
    \prelest = 1+ \sum_{i,j \in \Omega_{\fresnetact}} \underbrace{-\text{ReLU}(\fresnetact_{i,j})}_{\fcam}\,.
\end{equation}
Since this only consists of operators that are sub-differentiable \wrt the inputs, we can readily implement this using \pytorch and train it end to end. As opposed to regular \acp{CAM} that are computed using unconstrained activation maps, the proposed formulation assures that the activations are in a physically plausibale range and hence, a direct interpretation as physical quantities is possible.

Our experiments showed that for successful training, a scaling of \fresnetact by $\lvert\Omega_\fresnetact\rvert$ is necessary to counter an exploding \ac{MSE} loss. Further, we note that \fcam contains a small constant bias after training. We remove that bias by subtracting the median of multiple \fcam computed from module images with $\prel \approx 1$ from all \fcam.

The results are shown in~\cref{fig:cam} and additional examples can be found in~\cref{app:additional-cam}. It turns out that the resulting \acp{CAM} mostly highlight the fractures, indicating that the network is able to learn physically relevant features and that fractures are the main source of power loss as opposed to cracks.

It becomes apperent that, in many examples, the amount of power loss per cell predicted by the network is roughly proportional to the amount of inactive area. Although this might appear obvious at first, this is not a trivial finding, because there are various types of cracks and their relative position is important as well. For example, it is known that the current and with the the power of a string is limited by the worst cell, since all cells are connected in series. However, at least for this dataset, the network reveals a linear relationship between the amount of inactive area and module power. To verify this, we compute the size of inactive area for every module in the dataset by thresholding and calculate the power loss proportional to the inactive area\addition[label=ch012,ref=co022]{, which is similar to the thresholding approach by \etal{Schneller}~\cite{schneller2018electroluminescence}}. Using this approach, the module power is predicted with a \ac{MAE} of \SI[separate-uncertainty=true, multi-part-units=single]{8.5(21)}{\wattpeak}, which is on par with the baseline results reported in~\cref{fig:boxplots-mae-rmse-all}. \addition[label=ch013,ref=co022]{ It is possible that those results could be improved even further by using more than a single threshold, because this would allow the model to take variations in the shunt resistance into account.}

\subsection{Generalization to unseen Data}\label{subsec:ex-dl-generalization}

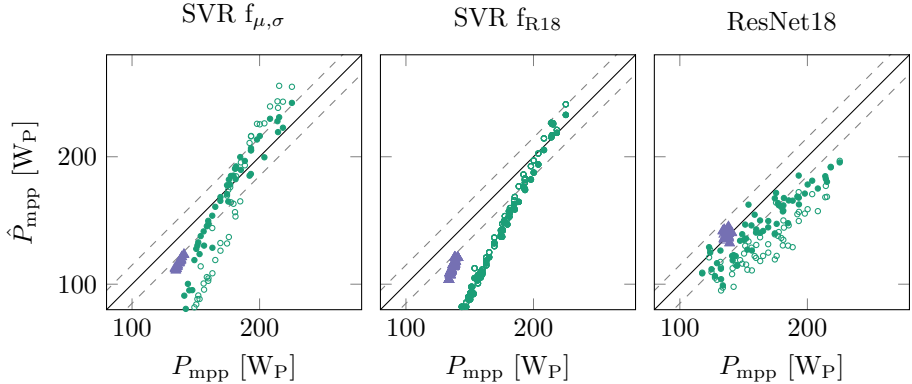
\begin{figure*}[t]
    \pgfplotsset{
        compat=newest,
        every axis/.append style={
            scatterplotnew style,
            xtick={100,200},
            ytick={100,200},
            title style={align=center, font=\linespread{0.8}\selectfont},
        }
    }

    \tikzsetnextfilename{scatter_generalization}%
    \begin{tikzpicture}
        \begin{groupplot}[
            group style={
                group size=3 by 1,
                vertical sep=1.5cm,
                ylabels at=edge left,
                xlabels at=edge bottom,
                yticklabels at=edge left,
                horizontal sep=.25cm,
            },
            width=1/2.3*\textwidth-8.5pt,
            height=1/2.3*\textwidth-8.5pt,
        ]

            \nextgroupplot[
                title={\ac{SVR}~\fmstd},
            ]
                \addplot [
                    scatter samples,
                ]table [x=peak_power, y=predicted_power, meta=source_key, col sep=comma]{csv_results/041_check_generalization/cv_mean_std_test_errors_0.csv};
                \scattercommon
            
            \nextgroupplot[
                title={\ac{SVR}~\fresnet},
                legend columns=-1,
                legend to name=scatter-legend-generalization,
                legend entries={indoor,on-site,\typee,\typeh,high current,low current},
                legend style={/tikz/every even column/.append style={column sep=0.25cm}},
            ]

                \addlegendimage{only marks,mark=o,black};
                \addlegendimage{only marks,mark=triangle,black};
                \addlegendimage{only marks,mark=*,colore3};
                \addlegendimage{only marks,mark=*,colorh2};
                \addlegendimage{only marks,mark=o,black};
                \addlegendimage{only marks,mark=*,black};

                \addplot [
                    scatter samples,
                ]table [x=peak_power, y=predicted_power, meta=source_key, col sep=comma]{csv_results/041_check_generalization/cv_pretrained_test_errors_all.csv};
                \scattercommon

            \nextgroupplot[
                title={\resneta},
            ]
            \addplot [
                scatter samples,
            ]table [x=peak_power, y=predicted_power, meta=source_key, col sep=comma]{csv_results/041_check_generalization/fold_0.csv};
            \scattercommon
            
        \end{groupplot}
    \end{tikzpicture}

    \tikzexternaldisable
    \caption{Distribution of estimation errors for the generalization experiment. For the samples, we distinguish between indoor~\ref{pgfplots:mark-indoor} and on-site \ref{pgfplots:mark-outdoor} measurements by marker shape. Further, we distinguish between module type \typee~\ref{pgfplots:mark-typee} and \typeh~\ref{pgfplots:mark-typeh} by marker color. Finally, we differentiate measurements taken at high current~\ref{pgfplots:mark-highcurrent} or low current~\ref{pgfplots:mark-lowcurrent} by marker filledness. For better visualization, the ideal regression line is shown~\ref{pgfplots:line-ideal}. Furthermore, \ref{pgfplots:line-dashed} indicates the \SI[separate-uncertainty=true, multi-part-units=single]{0(15)}{\wattpeak} isoline.}
    \tikzexternalenable
    \label{fig:scatter-generalization}
\end{figure*}

\begin{table}[t]
    \centering
    \csvloop{
        file=csv_results/041_check_generalization/summary.csv,
        head to column names,
        before reading=\sisetup{table-format=2.1,round-mode=places,round-precision=1,table-number-alignment=left},
        tabular={lS[table-format=2.1,separate-uncertainty=true,table-figures-uncertainty=3,table-align-uncertainty=true]S[table-format=2.1,table-number-alignment=left]S[separate-uncertainty=true,table-figures-uncertainty=4,table-align-uncertainty=true]S},
        table head={\toprule \linespread{0.1} & \text{MAE}~[\si{\percent}] & \text{RMSE}~[\si{\percent}] & \text{MAE}~[\si{\wattpeak}] & \text{RMSE}~[\si{\wattpeak}] \\ \midrule},
        command=\texmodel & \mae(\uncertainty) & \rmse & \maeW(\uncertaintyW) & \rmseW,
        table foot=\bottomrule
    }
    \caption{Quantitative results of the generalization experiment. We show the \ac{MAE} and the \ac{RMSE} as well as the standard deviation of errors. Finally, we explicitly show the relative error, as used in training as well as the absolute error in [\si{\wattpeak}] for better interpretability.}
    \label{tab:generalization-results}
\end{table}

Finally, we aim to assess, if the method generalizes well to module types that have not been used during training. This experiment is conducted using the \resneta, since this performed best in the previous experiments. In addition, we include the \ac{SVR}-based methods for reference. We train the model on samples \typet using the hyperparameters that have been found in the first fold of the \ac{CV} and test on samples \typee and \typeh. Note that the module types differ in their physical properties. For example, \typet and \typeh consist of $10 \times 6$ polycrystalline cells, of which each has an edge length of $6$~inch, whereas \typee consists of $12 \times 6$ cells with an edge length of $5$-inch. Furthermore, they also differ in their nominal power, which is given as \SI{230}{\wattpeak} for \typet, \SI{170}{\wattpeak} for \typee and \SI{240}{\wattpeak} for \typeh. However, all types have in common that their cells are arranged in $3$ substrings that are connected in parallel and include a bypass diode for every substring. 

The results are summarized by ~\cref{fig:scatter-generalization,tab:generalization-results}. It turns out that the performance degrades in this setting. For example, the performance of \resneta is lowered to \resultsgeneralizationmae{resnet18} (as opposed to \resultsmae{resnet18} when training on the whole dataset), while the performance drop for the \ac{SVR}-based methods is even larger.

From~\cref{fig:scatter-generalization}, it becomes apparent that, especially the \resneta generalizes well to \typee, although it has never been trained on monocrystalline modules. However, it does not generalize well to \typeh. This result is in line with the results reported in~\cref{tab:subset-results}, where it turns out that the model trained on all subsets performs worst on \typeh as well.

We conclude that the dataset bias, which is already observed in \cref{fig:data-tsne-pretrained-subsets}, limits the generalization ability of the method to unseen module types. This problem could most likely be solved by using a larger dataset that covers a greater variety of module types.

\section{Conclusion and Future Work}\label{sec:conclusion}

In this work, we propose a novel method to assess the power of individual \acs{PV} modules using a single \acl{EL} measurement only. We combine classical image processing methods for segmentation of multiple modules with a \acl{DL}-based prediction of the \acl{MP}. We find that our method is capable to predict the \acl{MP} with an average \acs{MAE} of \resultsmaew{resnet18} in a \acl{CV}. This is already close to the measurement error of the system, which is specified as \SI{2}{\percent} (indoor) and \SI{5}{\percent} (on-site), resulting in a lower bound to the \acs{MAE} of \SI{4.6}{\wattpeak}/\SI{11.5}{\wattpeak}. \addition[ref=co024,label=ch043]{Although the choice of a \resneta and the reported hyperparameters are specific to the dataset used in this work, we are confident that they hold for other datasets as well, since previous experiments on defect recognition lead to similar conclusions~\cite{mayr2019weakly}.}

\addition[ref=co023,label=ch042]{For this work, we use a dataset of \num{719} \ac{EL} measurements including three module types. However, we experimentally show that the model does not generalize well to unseen module types. This cannot be solved by data augmentation, since this is most likely caused by differing module properties like busbar configurations. We are confident, that the generalization can be improved by training on a larger dataset comprising more different types of modules. This is because a greater variety in the training data would help the model to learn the most important features for power prediction, rather than overfitting on certain defect types that only occur for a specific module type.} \addition[ref=co006,label=ch048]{We point out that the dataset is selected such that the modules have a high shunt resistance. Other defect types like potential induced degradation are not taken into consideration. As a result, the trained models are restricted to this particular type of defects.} \addition[ref=co024]{However, we are confident that the method can be applied to other defect types as well, as long as they are visible in \ac{EL} measurements and a comprehensive training dataset is available.}

\deletion[ref=co024]{Since there are neither standard datasets nor existing methods for comparison, a proper evaluation of the method turned out challenging. We solve this issue and set up a baseline method using \acs{SVR} regression of the \acl{MP} from simple features. This results in a \acs{MAE} of, showing that using \acl{DL} is beneficial for the task at hand.}

We transfer the concept of \aclp{CAM} to the regression task and use it to calculate the power loss per cell jointly with the power loss of the overall module. Since the per cell power loss has never been used during training, this approach allows to quantify the impact of individual defects on the overall power output and does not require any additional measurements or finite element analysis. \addition[ref=co024]{Our experiments show that the model can automatically determine the defect types that are most relevant to calculate the power loss of the module. By means of the thresholding experiment, we show that this quantification of power losses can help to design simpler methods that only perform slightly worse, compared to the \ac{DL}-based approach.}

Furthermore, we compile and release the dataset consisting of \num{719} \acs{EL} measurements that have been aquired under varying conditions \addition[ref=co031,label=ch046]{as well as measurements of the \pmpp at standard test conditions}. This includes indoor and on-site measurements as well as measurements from load cycle experiments. As some of the on-site measurements show multiple modules at the same time, we propose a simple yet robust preprocessing pipeline to detect and segment module instances. We evaluate the detection of modules on a separate testset and find that it results in a detection rate of $F_1 = 0.94$. \addition[ref=co027,label=ch045]{This detection pipeline is currently designed to detect modules that do not have a defective substring. Modules with a defective substring result in two separate detections, since the parts are not conneceted in the binarized measurement. An extension of the method for such cases is subject to future works.}

\deletion[ref=co024]{Since the dataset is of limited size, we conduct a $5$-fold \ac{CV} to increase the significance of  quantitave results. Here, we also perform a hyperparameter search for the batch size, the weight decay and the learning rate in every fold of the \ac{CV} using \num{250} iterations. Further, we compare \mobilenet, \resneta and \resnetb architectures. This results in \num{3750} trained models in total.}

\deletion[ref=co024]{For the hyperparameter optimization, we decided to focus on only a few parameters that turned out to have the largest impact on the result in our experiments. This is to avoid combinatorial explosion of the search space. However, there are many more parameters that could influence the result. Therefore, we conduct initial experiments to determine the optimal image size as well as normalization scheme.}

For future works, we think that it might be interesting to consider architectures that operate on cell level, because they could take the electrical connections into account, which is not considered in this work. As shown in~\cref{subsec:ex-dl-cam}, the model mainly focusses on fractures, which is consistent to physical considerations \addition[ref=co002]{and prior works}. However, the impact of a fracture on the \ac{MP} is dependent on the overall conductivity of the substring. This relationship could be learned from the data using an architecture that determines the fraction of inactive area per cell and subsequently combines this information into a final estimate. Further, the generalization gap shown in~\cref{fig:scatter-generalization} deserves further investigation.

We are confident that this work contributes to an automated and contactless assessment of large \ac{PV} power plants. By releasing the data and code, we aim to allow other researchers to reproduce our results and to push the field forward.

\subsubsection*{Acknowledgements}

We gratefully acknowledge the BMWi as well as the IBC SOLAR AG for financial funding of the project iPV4.0 (FKZ: 0324286) and the German Federal Ministry for Economic Affairs and Energy (BMWi) for financial funding of the project COSIMA (FKZ: 032429A). Furthermore, we acknowledge the PV-Tera grant by the Bavarian State Government (No. 44-6521a/20/5). Further, we thank Allianz Risk Consulting GmbH / Allianz Zentrum für Technik (AZT) for providing us with a large number of \ac{PV} modules.

\clearpage

\bibliography{mybibfile}

\clearpage
\begin{appendices}
    \crefalias{section}{appendix}

    \section{Detection results\label{app:detection-results}}

    \begin{figure}[h]
        \centering
        \tikzsetnextfilename{preproc_pr}
        \begin{tikzpicture}[]
            \begin{axis}[
                xlabel={\pretiou},
                ylabel={precision / recall~[\%]},
                legend pos=south west,
                width=0.75\textwidth,
                height=0.4\textwidth,
            ]
                \addplot+ table [x=iou, y expr=100*\thisrow{precision}, col sep=comma]{csv_results/preprocessing/results_iou.csv};
                \addplot+ table [x=iou, y expr=100*\thisrow{recall}, col sep=comma]{csv_results/preprocessing/results_iou.csv};
                \legend{precision (AUC: $0.94$),recall (AUC: $0.90$)};
            \end{axis}
        \end{tikzpicture}
        \caption{Detection performance of the proposed method. We compute the \ac{IoU} of predicted and ground truth boxes. Then, we state a prediction as accepted, if the \ac{IoU} with the ground truth box is greater than \pretiou. Finally, we compute precision and recall at varying thresholds \pretiou. It turns out that the method works very well up to $\pretiou = 0.85$ and slightly drops in precision and recall beyond that threshold.}
        \label{fig:preproc-pr}
    \end{figure}
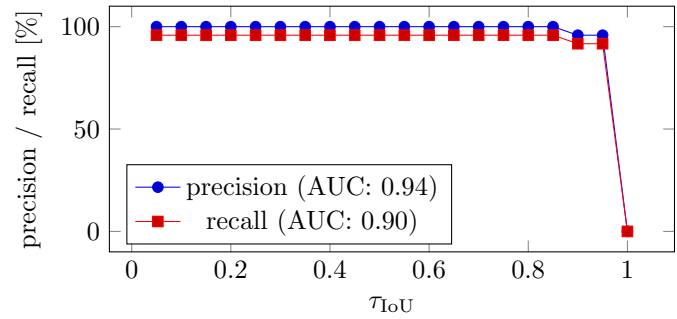

    \clearpage

    \section{Additional \acp{CAM}}\label{app:additional-cam}

    \begin{figure}[h]
        \begin{subfigure}{\textwidth}%
            \centering%
            \tikzsetnextfilename{cam2}%
            \begin{tikzpicture}[
            ]
                \begin{axis}[
                    width=0.88\textwidth,
                    height=0.58\textwidth,
                    hide axis,
                    y dir=reverse,
                    colorbar,
                    colormap/viridis,
                    point meta min=0,
                    point meta max=100,
                    colorbar style={
                        ytick={0, 25, 50, 75, 100},
                        title={[\%]},
                    },
                    enlargelimits=false,
                ]
                    \addplot graphics[xmin=0, ymin=0, xmax=1333, ymax=800] {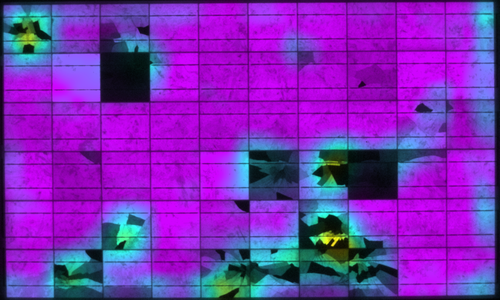};

                    \addplot [
                        mark=none,
                        draw=none,
                        nodes near coords={\tiny\num[round-mode=places, round-precision=1, tight-spacing=true]{\pgfplotspointmeta}},
                        nodes near coords align={anchor=center},
                        nodes near coords style={
                            rectangle,
                            fill=white,
                            fill opacity=0.7,
                            text opacity=1,
                            inner sep=2pt,
                            rounded corners=2pt,
                        },
                    ] 
                    table[
                        x=x,
                        y=y,
                        meta=v,
                        col sep=comma,
                        point meta=explicit symbolic
                        ] {csv_results/042_cam2/538.csv};

                \end{axis}
            \end{tikzpicture}
        \end{subfigure} \\
        \begin{subfigure}{\textwidth}%
            \centering%
            \tikzsetnextfilename{cam3}%
            \begin{tikzpicture}[
            ]
                \begin{axis}[
                    width=0.88\textwidth,
                    height=0.58\textwidth,
                    hide axis,
                    y dir=reverse,
                    colorbar,
                    colormap/viridis,
                    point meta min=0,
                    point meta max=100,
                    colorbar style={
                        ytick={0, 25, 50, 75, 100},
                        title={[\%]},
                    },
                    enlargelimits=false,
                ]
                    \addplot graphics[xmin=0, ymin=0, xmax=1333, ymax=800] {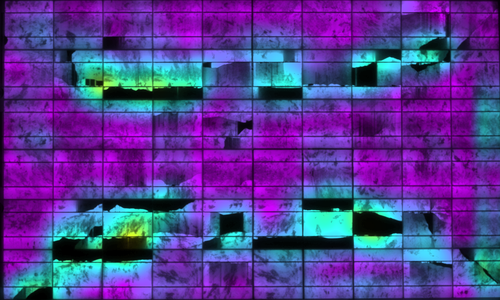};

                    \addplot [
                        mark=none,
                        draw=none,
                        nodes near coords={\tiny\num[round-mode=places, round-precision=1, tight-spacing=true]{\pgfplotspointmeta}},
                        nodes near coords align={anchor=center},
                        nodes near coords style={
                            rectangle,
                            fill=white,
                            fill opacity=0.7,
                            text opacity=1,
                            inner sep=2pt,
                            rounded corners=2pt,
                        },
                    ] 
                    table[
                        x=x,
                        y=y,
                        meta=v,
                        col sep=comma,
                        point meta=explicit symbolic
                        ] {csv_results/042_cam2/200.csv};

                \end{axis}
            \end{tikzpicture}
        \end{subfigure}

        \caption{Additional examples for \acp{CAM} used to compute the per cell power loss. }
        
    \end{figure}

    \clearpage

    \section{Data analysis}\label{app:data-analysis}

    \begin{figure}[h]
        \centering%
        \begin{NoHyper}\tikzexternaldisable\ref{scatter-all-legend}\tikzexternalenable\end{NoHyper}%
        \vspace{.3cm} \\%
        \subcaptionbox{}{%
            \tikzsetnextfilename{explore_data_pretrained_embeddings_sets}%
            \begin{tikzpicture}%
                \begin{axis}[
                    scatterclasses style,
                    only marks,
                    ticks=none,
                    width=4.5cm,
                    height=4cm,
                    scale only axis=true,
                ]

                    \addplot [
                        scatter samples
                    ] table [x=x, y=y, meta=source_key, col sep=comma]{csv_results/005_explore_data/embeddings_all.csv};

                \end{axis}
            \end{tikzpicture}%
            \label{fig:data-tsne-pretrained-subsets}%
        }%
        \hfill%
        \subcaptionbox{}{
            \tikzsetnextfilename{explore_data_pretrained_embeddings}%
            \begin{tikzpicture}%
                \begin{axis}[
                    only marks,
                    colorbar,
                    legend style={
                        legend pos=south east
                    },
                    ticks=none,
                    colorbar style={
                        title=\prel~[\si{\percent}]
                    },
                    width=4.5cm,
                    height=4cm,
                    scale only axis=true,
                ]

                    \addplot+[scatter, scatter src=explicit, mark=*, mark options={scale=0.5}, point meta={100*\thisrow{relative_power}}]
                        table [x=x, y=y, col sep=comma]{csv_results/005_explore_data/embeddings_all.csv};
                \end{axis}
            \end{tikzpicture}%
            \label{fig:data-tsne-pretrained-power}%
        }%
        \caption{\Ac{t-SNE} visualization of \resneta embeddings.}
        \label{fig:data-tsne-pretrained}
    \end{figure}
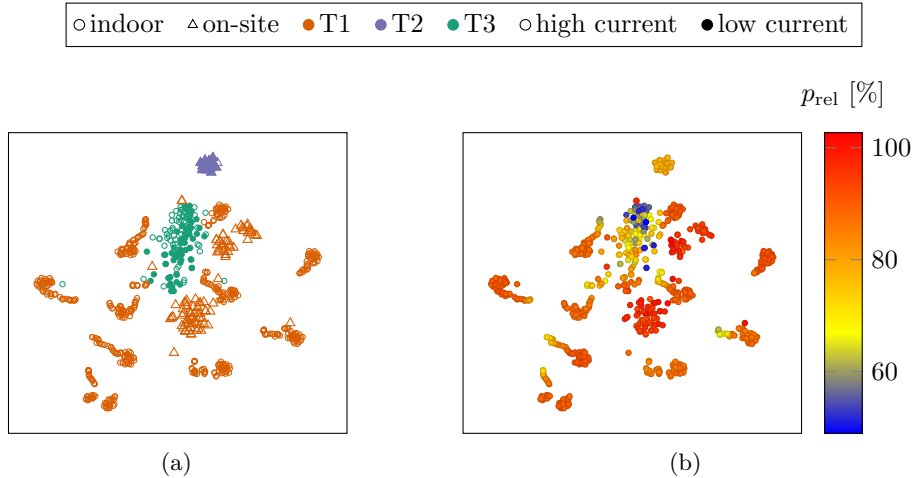
    
    In~\cref{fig:data-tsne-pretrained}, we visualize our data using \acf{t-SNE}~\cite{hinton2003stochastic} applied on the embeddings from a pretrained \resneta. We observe that there is a weak clustering according to \prel without any further training. This is in line with the results from an earlier publication on the same topic, where a pretrained \resneta already performed well on this task by only training the last fully connected layer~\cite{buerhop2019dlpower}. In addition, \cref{fig:data-tsne-pretrained} reveals that module instances from the load cycle measurements result in dense clusters. A further analysis of the data shows that there is a strong clustering regarding module types and measurement conditions in the \ac{t-SNE} visualization, as shown in~\cref{fig:data-tsne-pretrained-subsets}. We conclude that \prel is not the main mode of variation without further finetuning of the network. This observation is supported by the results obtained from using a pretrained \resneta followed by a \ac{SVR}, which we introduce as a baseline method~(\cref{subsec:pp-baseline}).

\end{appendices}

\end{document}